%% file: main.tex
\documentclass{article}

\input{preamble.tex}

\title{Efficient Statistics for Sparse Graphical Models from Truncated Samples}

\author[1]{Arnab Bhattacharyya\thanks{\texttt{arnabb@nus.edu.sg}. Supported by a National Research Foundation, Singapore Fellowship for AI (WBS R-252-000-B13-281) and an Amazon Research Award (WBS R-252-000-A61-720).}}
\author[1]{Rathin Desai\thanks{\texttt{desairathin18@gmail.com}. Supported by an Amazon Research Award (WBS R-252-000-A61-720).}}
\author[2]{Sai Ganesh Nagarajan\thanks{\texttt{sai\_nagarajan@mymail.sutd.edu.sg}. Supported by SUTD President's Graduate Fellowship (SUTD-PGF).}}
\author[3]{Ioannis Panageas\thanks{\texttt{ioannis@sutd.edu.sg}. Supported by NRF-NRFFAI1-2019-0003, SRG ISTD 2018 136 and ANR NRF 0095 ALIAS.}}
\affil[1]{{ School of Computing\\ National University of Singapore}}
\affil[2]{{Engineering Systems and Design\\ Singapore University of Technology and Design}}
\affil[3]{{Information Systems\\ Singapore University of Technology and Design}}

\begin{document}

\maketitle

\begin{abstract}%
In this paper, we study high-dimensional estimation from truncated samples. We focus on two fundamental and classical problems: (i) inference of sparse Gaussian graphical models and (ii) support recovery of sparse linear  models.
\begin{itemize}
\item[(i)]
For Gaussian graphical models, suppose $d$-dimensional samples $\vec{x}$ are generated from a Gaussian $N(\vec{\mu},\vec{\Sigma})$ and observed only if they belong to a subset $S \subseteq \R^d$. We show that  $\vec{\mu}$ and $\vec{\Sigma}$ can be estimated with error $\epsilon$ in the Frobenius norm, using\footnote{$\textrm{nz}(A)$ denotes the number of non-zero entries of matrix $A$.} $\tilde{O}\left(\frac{\textrm{nz}(\vec{\Sigma}^{-1})}{\epsilon^2}\right)$ samples from a truncated $\mathcal{N}(\vec{\mu},\vec{\Sigma})$ and having access to a membership oracle for $S$. The set $S$ is assumed to have non-trivial measure under the unknown distribution but is otherwise arbitrary.
\item[(ii)]
For sparse linear regression, suppose samples $(\vec{x},y)$ are generated where $y =  \vec{x}^\top{\vec{\Omega}^*} + \mathcal{N}(0,1)$ and $(\vec{x}, y)$ is seen only if $y$ belongs to a truncation set $S \subseteq \mathbb{R}$. We consider the case that $\vec{\Omega}^*$ is sparse with a support set of size $k$. Our main result is to establish precise conditions on the problem dimension $d$, the support size $k$, the number of observations $n$, and properties of the samples and the truncation that are sufficient to recover the support of $\vec{\Omega}^*$. Specifically, we show that under some mild assumptions, only $O(k^2 \log d)$ samples are needed to estimate $\vec{\Omega}^*$ in the $\ell_\infty$-norm up to a bounded error.
\end{itemize}
For both problems, our estimator minimizes the sum of the finite population negative log-likelihood function and an $\ell_1$-regularization term. 
\end{abstract}%


\input{intro}
\input{prelims}

\input{frobeniusv3}

\input{linreg}
\input{future}

\section*{Acknowledgements}
AB would like to thank Chiranjib Bhattacharyya for motivating us to study the support recovery problem with truncated samples.
\ignore{\newpage

\section*{Broader Impact}
Research in parameter estimation from truncated or corrupted data is a fundamental topic in Statistics and Machine Learning more broadly. Our work is theoretical and we do not believe it has any ethical or future societal consequences.}
\bibliographystyle{plain}
\bibliography{papers}

\appendix
\input{ell2}

\end{document}

%% file: preamble.tex
\usepackage[utf8]{inputenc} 
\usepackage[T1]{fontenc}    
\usepackage{url}            
\usepackage{booktabs}       
\usepackage{nicefrac}       
\usepackage{microtype}      
\usepackage{fullpage}
\usepackage{xr}

\usepackage{authblk}

\makeatletter
\newcommand{\opnorm}{\@ifstar\@opnorms\@opnorm}
\newcommand{\@opnorms}[1]{%
  \left|\mkern-1.5mu\left|\mkern-1.5mu\left|
   #1
  \right|\mkern-1.5mu\right|\mkern-1.5mu\right|
}
\newcommand{\@opnorm}[2][]{%
  \mathopen{#1|\mkern-1.5mu#1|\mkern-1.5mu#1|}
  #2
  \mathclose{#1|\mkern-1.5mu#1|\mkern-1.5mu#1|}
}
\makeatother

\usepackage{amsmath, amssymb, amsfonts}
\usepackage{ifthen}
\usepackage{comment}
\usepackage[unicode,psdextra,colorlinks=true,citecolor=blue,bookmarksnumbered=true,linkcolor=green!40!black]{hyperref}
\usepackage{pgf}
\usepackage{tikz}
\usetikzlibrary{arrows,automata}
\usepackage{bm}

\usepackage{graphicx}
\usepackage{float}
\usepackage{caption}
\usepackage{subcaption}

\usepackage{amsthm}
\usepackage{multirow, multicol}
\newtheorem{theorem}{Theorem}[section]

\newtheorem{claim}[theorem]{Claim}

\newtheorem{lemma}[theorem]{Lemma}

\newtheorem{assumption}[theorem]{Assumption}

\newtheorem{definition}[theorem]{Definition}

\numberwithin{equation}{section}

\newcommand{\ignore}[1]{}

\newcommand{\defeq}{\stackrel{\mathrm{def}}=}

\newcommand{\norm}[1]{\left\| #1 \right\|}

\renewcommand{\vec}[1]{\bm{\mathrm{#1}}}

\definecolor{DSred}{rgb}{1,0,0}

\renewcommand{\leq}{\leqslant}
\renewcommand{\geq}{\geqslant}

\renewcommand{\le}{\leqslant}
\renewcommand{\epsilon}{\varepsilon}
\newcommand{\eps}{\epsilon}

\newcommand{\R}{\mathbb{R}}

\newcommand{\EE}{\mathbb{E}}
\newcommand{\PP}{\mathbb{P}}

\DeclareMathOperator*{\E}{\EE}

\renewcommand{\Pr}{\PP}
\DeclareMathOperator*{\Cov}{\mathbf{Cov}}



%% file: intro.tex

\section{Introduction}

Sparse high-dimensional models are a mainstay of modern statistics and machine learning. In this work, we consider two different sparse linear models that have been the subject of intensive study.
\begin{itemize}
\item{\textbf {Sparse Gaussian Graphical Models.}}
Graphical models are used to represent the probabilistic relationships between a collection of variables. These models are used in a huge number of different domains, such as statistical physics, computational biology, finance, and machine learning; the books \cite{Laur96, Mez08, WaiJo08, KoFi10} give an indication of the breadth of this area. We focus on Gaussian graphical models in which the $d$ variables $X_1, \dots, X_d$ are distributed according to a $d$-dimensional Gaussian. Specifically, the distribution is described in terms of a density function $p(\vec{X})$ where $\vec{X}=(X_1, \dots, X_d)$ and
$$p(\vec{X}) = (2\pi)^{-d/2}\cdot (\det \vec{\Sigma})^{-1/2} \exp\left(-\frac12 (\vec{X}-\vec{\mu})^\top \vec{\Sigma}^{-1} (\vec{X}-\vec{\mu})\right).$$
Here, $\vec{\mu}$ and $\vec{\Sigma}$ correspond to the mean and variance of the distribution respectively.

It is convenient to reparametrize the density function in terms of the inverse covariance matrix or the {\em precision matrix}, $\vec{\Theta} = \vec{\Sigma}^{-1}$:
$$p(\vec{X}) = (2\pi)^{-d/2}\cdot \exp\left(\vec{\mu}^T\vec{\Theta}\vec{X} - \frac12 \vec{X}^\top\vec{\Theta}\vec{X}-\frac12 \vec{\mu}^\top \vec{\Theta}\vec{\mu} + \frac12 \log \det(\vec{\Theta})\right)$$
Note that the exponent is a quadratic polynomial in which the coefficient of $X_iX_j$ is $\Theta_{i,j}$. The symmetric matrix $\vec{\Theta}$ naturally defines an undirected graph $G$ on $d$ vertices in which $(i,j)\in E(G)$ iff $\Theta_{i,j} \neq 0$. The graph $G$ also admits a very nice probabilistic interpretation: $X_i$ and $X_j$ are independent conditioned on all other variables if and only if $\Theta_{i,j}=0$. Thus, for natural systems, it is quite reasonable to assume that the degree of each node in $G$ is small, as this corresponds to assuming that each variable is ``directly'' dependent on a small number of variables. Note that even if $\vec{\Theta}$ is sparse, $\vec{\Sigma}$ could be dense; in fact, in many typical systems, any pair of variables is correlated even though they are not directly dependent.

The problem of learning sparse high-dimensional Gaussian graphical models (in terms of the precision matrix) has a rich history. Popular approaches include the graphical Lasso \cite{Fri07, Yuan07, Ban08, Dasp08, Roth08, Ravi11}, neighborhood-based methods \cite{Bes74, Mein06, Wai09}, and CLIME \cite{Cai11} which have been proved to work under different sets of assumptions.

\item{\textbf{Sparse Linear Regression.}}
A fundamental problem in data science is to solve the following inverse problem. Given pairs $(\vec{x}_1, y_1), \dots, (\vec{x}_n,y_n)  \in \R^d \times \R$, find the ``best'' choice of $\vec{\Omega} \in \R^d$ so that $y_i - \vec{x}_i^\top \vec{\Omega}$ is small in some norm. It is natural to want $\vec{\Omega}$ to be sparse so that the prediction can be made based on a small number of variables.

Consider the model $y = \vec{x}^\top {\vec{\Omega}^*} + \epsilon$ where $\epsilon$ is a Gaussian random variable and $\vec{\Omega}^*$ is a sparse vector. There has been a huge amount of work on this problem. In the high-dimensional setting, a very popular approach is using $\ell_1$-regularization, leading to the Lasso algorithm \cite{Tib94}. By now, we have an almost complete understanding of the necessary and sufficient conditions needed for Lasso to recover $\vec{\Omega}^*$; see the discussion and references in Chapter 7 of \cite{Wbook}.
\end{itemize}

In our work, we study the above two problems in the setting where the samples are subject to {\em truncation}. Truncation is also a classic challenge in statistics, occurring whenever the observation process is dependent on the drawn sample. Following early work by Galton \cite{Gal98}, there has been a sustained history of research on truncated distributions, in particular, truncated Gaussians (see the citations in \cite{Coh16}) and truncated linear regression \cite{Tob58, Ame73, HW77, Breen96}. We pick up the thread at \cite{Das18}  ~who developed a computationally and statistically efficient algorithm to learn a multivariate Gaussian given truncated samples and assuming that the truncation set is known. A follow-up work, \cite{Das19}, extended the analysis to the linear regression problem where only those samples $(\vec{x},y)$ are seen in which $y \in S$, the truncation set.

To the best of our knowledge, ours is the first work that examines the problems of learning sparse Gaussian graphical models and linear models with truncated samples. We state our results next.

\paragraph{Statement of the results}
The first contribution of the paper is the following theorem on learning Gaussian graphical models up to small Frobenius norm error. The sampling process is that samples from an unknown $d$-variate Gaussian are only revealed if they belong to a subset $S \subseteq \R^d$; otherwise, the samples are completely hidden.
\begin{theorem}[Frobenius norm]\label{thm:main1}
Suppose that we are given oracle access to a measurable set $S$, so that $\int_S \mathcal{N}(\vec{x};\vec{\mu}^*, \vec{\Sigma}^*) d\vec{x} = \alpha >0$ for some d-variate $\mathcal{N}(\vec{\mu}^*,\vec{\Sigma}^*)$ and constant\footnote{Think of $\alpha$ like $1\%$.} $\alpha > 0$. There exists an estimator $\tilde{\vec{\mu}},\tilde{\vec{\Sigma}}$ that uses $\tilde{O}\left(\frac{\textrm{nz}(\vec{\Sigma}^{* -1})}{\epsilon^2}\right)$ samples from the truncated distribution $\mathcal{N}(\vec{\mu}^*,\vec{\Sigma}^*,S)$  so that with probability at least $99\%$
\[
\norm{I-\vec{\Sigma}^{* -1/2}\tilde{\vec{\Sigma}}\vec{\Sigma}^{* -1/2}}_F \leq \epsilon \textrm{ and }\norm{\vec{\Sigma}^{* -1/2}(\vec{\mu}^*-\tilde{\vec{\mu}})}_2 \leq \epsilon.
\]
\end{theorem}
The second contribution of the paper solves the variable selection problem for linear models, under certain assumptions. The sampling process is as follows: each $\vec{x}^{(i)} \in \R^d$, a row vector of $d$ covariates, is picked arbitrarily, and the value $y^{(i)} = \vec{x}^{(i)} {\vec{\Omega}^*}  + \epsilon^{(i)}$ is revealed only if $y_i \in S$. Here, $\eps^{(i)} \sim \mathcal{N}(0,1)$, the standard normal distribution.
\begin{theorem}[Linear regression, informal]\label{thm:main2}

Suppose that we are given oracle access to a measurable set $S$. Let $\vec{X}$ denote the design matrix whose rows are $\vec{x}^{(1)}, \dots, \vec{x}^{(n)} \in \R^d$. Let $K$ denote the unknown support of
 $\vec{\Omega}^*$, and let $k = |K|$. Assume that:

\begin{itemize}

\item[(i)] (\textbf{Survival probability})
For each observed $\vec{x}^{(i)}$, the probability that $\vec{x}^{(i)}{\vec{\Omega}^*}  + \eps_i$ survives the truncation is not too small.

\item[(ii)] (\textbf{Minimum eigenvalue}) The vector $\vec{\Omega}^*$ is identifiable if its support $K$ was known a priori.

\item[(iii)] (\textbf{Mutual incoherence}) Covariates not in the support set $K$ form columns in $\vec{X}$ that are approximately orthogonal to the space spanned by the columns corresponding to $K$.

\item[(iv)] (\textbf{Normalization}) Each entry of $\vec{X}$ is small in magnitude.

\end{itemize}

Then, with only $n=O(k^2 \log d)$ samples $(\vec{x}^{(i)}, y_i)$ from the truncated distribution, one can recover a vector $\hat{\vec{\Omega}}$ such that with high probability:

\begin{itemize}

\item[(a)] The support of $\hat{\vec{\Omega}}$ is contained in $K$.

\item[(b)] If for some $j \in K$, ${\Omega}^*_j$ is larger than a threshold $\tau$ (which depends on the problem parameters but not $d$), then ${\hat{\Omega}}_j \neq 0$.

\end{itemize}

\end{theorem}

We can also prove guarantees for recovering $\vec{\Omega}^*$ to within bounded $\ell_2$-error; see Appendix \ref{sec:ell2}. In fact in this setting, the error can be driven to any $\eps>0$ using roughly $\tilde{O}(k^2 \log(d) \eps^{-2})$ samples with assumptions similar to that above. 

\input{techniques.tex} 

\paragraph{Other related works}
Our work comes under the purview of robust statistics where the body of work relating to \cite{diakonikolas2017robust,diakonikolas2017robustly,diakonikolas2016robust,lai2016agnostic,charikar2016learning} provided guarantees for computationally efficient robust estimators in the presence of corruptions of an $\epsilon$ fraction of the data, when the samples are drawn from a multivariate Gaussian distribution. In addition, \cite{diakonikolas2017statistical} provide statistical query lower bounds on estimation problems related to multivariate Gaussians such as learning mixtures of high dimensional Gaussians. These works generally talk about the seemingly inherent trade-off between increasing the sample complexity for computational tractability. As a result, an important assumption about the underlying problem or the statistical model is that of sparsity. Aside from the works related to estimation in sparse models in classical statistics such as sparse linear regression (LASSO) and sparse PCA \cite{zou2006sparse} to mention a few, there is a line of work related to robust estimation in sparse models, such as robust sparse mean estimation when the covariance matrix is identity and then detection of rank 1 sparse shifts of high dimensional covariances of Gaussian distributions when the mean is zero, using the spiked covariance model as studied in \cite{li2017robust,pmlr-v65-balakrishnan17a}. 

%% file: techniques.tex

\paragraph{Our techniques}
We first discuss the ideas behind Theorem \ref{thm:main1}.
In \cite{Das18}, it was shown that using $n = \tilde{O}\left(\frac{d^2}{\epsilon^2}\right)$ samples from a $d$-variate truncated Gaussian distribution with truncation set $S$ of measure some constant $\alpha>0$, the mean $\vec{\mu}^*$ and the covariance $\vec{\Sigma}^*$ of the untruncated distribution can be estimated with $\epsilon$ error in $\ell_2$ and Frobenius norm respectively. The crux of their proof involves proving that the infinite population negative log-likelihood is $\kappa$-strongly convex in a neighborhood $U$ ($U \subseteq \mathcal{S}_{d\times d} \times \mathbb{R}^d$) of the true parameters where the radius of $U$ and $\kappa$ are functions of $\alpha$\footnote{Think of the radius $r$ as $O\left(\frac{\log (1/\alpha)}{\alpha^2}\right)$ and $\kappa$ to be $O(\alpha^{cr^5})$ where $c$ some constant. $U$ is a subset of $\mathcal{S}_{d\times d}\times \mathbb{R}^d$ where $\mathcal{S}_{d\times d}$ denotes the symmetric matrices of size $d\times d$.}. Moreover, they run projected SGD with an efficient projection procedure in that neighborhood $U$. SGD requires a sample from the true truncated distribution in every iteration, so the sample complexity of this approach is at least as much as the number of iterations of SGD. Due to variance reasons, for SGD to converge, the number of samples needed is $\Omega\left(\frac{d^2}{\epsilon^2}\right)$.

To improve up on their sample complexity, our estimator is the minimizer of a different function - denoted by $L_n$ - which is the \textit{finite} population negative log-likelihood plus a regularization term (see Equation (\ref{eq:newfunction})). The regularization term is the sum of the absolute values of the entries of the precision matrix (excluding the diagonal entries). This approach is the well-known Graphical Lasso.

One first easy observation is that the finite population negative log-likelihood and the infinite population negative log-likelihood have the same Hessian (thus same convexity properties, see Equation (\ref{eq:same})). Moreover, since the extra regularization term does not change the convexity properties of the finite population negative log-likelihood, we get for free from \cite{Das18} that the function $L_n$ is $\kappa$-strongly convex in a neighborhood $U$ of the true parameters (same $\kappa$ and $U$ as before). The crucial part now is that for the Lasso approach to work, we need that the empirical mean and the empirical covariance (from the truncated distribution) is close in $\ell_{\infty}$ and max-norm respectively (and not in $\ell_2$ and Frobenius norm). The only requirement for the proof to go through is that the number of samples gives the statistical guarantee for Lasso to work (see Lemma \ref{lem:mainbig}). 

For the support recovery problem in the sparse linear model with truncated samples, we again consider the Lasso objective, i.e., the sum of the finite population negative log-likelihood plus $\lambda \|\vec{\Omega}\|_1$. This objective function is globally convex. Suppose we already know the support $K$ of $\vec{\Omega}^*$, the true $k$-sparse coefficient vector.  In this case, we can solve the Lasso objective restricted to the variables in $K$ and hope that it is strongly convex so that the minimum is unique. For the untruncated case, the minimum eigen	value assumption (Assumption (ii) in Theorem \ref{thm:main2}) implies global strong convexity. In the truncated case, we can only guarantee strong convexity in a neighborhood around $\vec{\Omega}^*$. By tuning the regularization parameter $\lambda$, we can ensure that the minimum of the restricted Lasso objective will be in this neighborhood, and hence, is uniquely defined. 

The main challenge in proving Theorem \ref{thm:main2} is to extend the above ideas to  when $K$ is not known. To this end, we use the {\em primal-dual witness method} that has proven very useful for studying many Lasso-type algorithms \cite{Wai09, OWJ08, JSRR10, CPW10, NW11, RWRY11, LST13, WLX14, XCL16}. We identify a strict dual feasibility condition that implies uniqueness of the Lasso solution and then demonstrate for a set of parameters that the condition holds. In contrast to the  untruncated case, we are not able to drive the $\ell_\infty$-error to zero as $n$ grows to infinity. Also, we require a stronger normalization on the entries of the design matrix. We leave  as an interesting open problem the question of overcoming these deficiencies in our analysis. 

%% file: prelims.tex
\section{Preliminaries} \label{sec:prelims}

\paragraph{Notation} We use bold faces to denote vectors and matrices. By $\vec{x}_{-j}$ we denote the vector $\vec{x}$ that involves all coordinates but $j$. We use $\textrm{vec}(\vec{A})$ to denote the standard vectorization of matrix $\vec{A}$. Moreover, we use $\norm{\textrm{vec}(\vec{A})}_{1,\textrm{off}}$ to denote the $\ell_1$ norm of $\textrm{vec}(\vec{A})$ by excluding the diagonal entries of matrix $A$ and $\textrm{nz}(\vec{A})$ for the number of non-zero entries of matrix $\vec{A}$. We denote by $\mathcal{S}_{d\times d}$ the set of symmetric matrices.

\paragraph{Norms} For a $d \times d$ matrix $\vec{A}$,
\begin{align*}
\norm{\vec{A}}_2 = \max_{\norm{\vec{x}}_2 = 1} \norm{\vec{Ax}}_2, \: \norm{\vec{A}}_{\infty} = \max_{j \in [n]} \sum_{i=1}^n |A_{ij}|, \: \norm{\vec{A}}_F = \sqrt{\sum_{i=1}^n \sum_{j=1}^n A_{ij}^2}.
\end{align*}
When $\vec{A}$ is a symmetric matrix we have that $\norm{\vec{A}}_2 \le \norm{\vec{A}}_{\infty} \le \norm{\vec{A}}_F \le \sqrt{n}\norm{\vec{A}}_2 \le \sqrt{n}\norm{\vec{A}}_\infty$. For a vector $\vec{x} \in \mathbb{R}^d$ we also have,
\begin{align*}
\norm{\vec{x}}_2 = \sqrt{\sum_{i=1}^d {x}_i^2}, \: \norm{\vec{x}}_{\infty} = \max_{j \in [d]} |{x}_j| , \: \norm{\vec{x}}_1 = \sum_{i=1}^d |{x}_i|.
\end{align*}
It holds that $\ell_{1}$ is the dual of $\ell_{\infty}$ and for $\vec{x},\vec{y} \in \mathbb{R}^d$ one can have $\vec{x}^T \vec{y} \leq \norm{\vec{x}}_1 \norm{\vec{y}}_{\infty}$ (H\"older's inequality).
\paragraph{Truncated Gaussian Distribution}
For a measurable set $S$ with parameters $\vec{\mu}, \vec{\Sigma}$, the density function for the truncated Gaussian distribution $\mathcal{N}(\vec{\mu},\vec{\Sigma}, S)$ with mean $\vec{\mu}$ and covariance $\vec{\Sigma}$ is defined as follows:
\begin{align}
\mathcal{N}(\vec{\mu},\vec{\Sigma},S;\vec{x})\defeq
{
	\begin{cases}
	\dfrac{\mathcal{N}(\vec{\mu},\vec{\Sigma};\vec{x})}{\int_{S}\mathcal{N}(\vec{\mu},\vec{\Sigma};\vec{x}) d\vec{x}} \;,\; \vec{x} \in S\\
	0                                            \;,\; \vec{x} \notin S
	\end{cases}
}
\end{align}
where
\[
\mathcal{N}(\vec{\mu},\vec{\Sigma};\vec{x}) \defeq \frac{1}{\sqrt{2\pi \textrm{det}(\vec{\Sigma})}}\exp\left(-\frac{1}{2} (\vec{x}-\vec{\mu})^T \vec{\Sigma}^{-1} (\vec{x}-\vec{\mu})\right).
\]
Throughout this paper, we will assume that the set $S$ can be accessed through a {\em membership oracle}.
\begin{definition}[Membership oracle] Let $S \subset \mathbb{R}^d$ be a measurable set. A membership oracle of $S$ is a function that given an arbitrary $\vec{x} \in \mathbb{R}^d$, it returns yes if it belongs to the set, otherwise no (i.e., it implements the indicator function of $S$). We assume oracle access to the indicator of $S$.
\end{definition}

\paragraph{Precision matrix and sparsity}
Let $G=(V,E)$ be an undirected graph with $V= [d]$. A random vector $\vec{x} \in \mathbb{R}^d$ is said to be distributed according to the (undirected) Gaussian Graphical model with graph $G$ if $\vec{X}$ has a multivariate Gaussian distribution $\mathcal{N}(\vec{\mu}, \vec{\Sigma})$ with
\begin{align}
\left(  \vec{\Sigma^{-1}} \right)_{ij} = 0 \; \; \forall \: (i,j) \notin E,
\end{align}
$\vec{\Sigma}^{-1}$ which we denote by $\vec{\Theta}$ is known as the precision matrix. In our results, the sample complexity depends on the number of non-zero entries of $\vec{\Sigma}^{-1}$, i.e., $\textrm{nz}(\vec{\Sigma}^{-1})$.

\paragraph{Strong Convexity}
\begin{lemma}[folklore, see p.309 of \cite{Wbook}]\label{lem:wainwright99} Suppose that a differentiable function $f : \mathbb{R}^d \to \mathbb{R}$ is $\kappa$-strongly convex in the sense that
\begin{equation}\label{eq:9.100a}
f(\vec{y}) \geq f(\vec{x}) + (\vec{y}-\vec{x})^T\nabla f(\vec{x}) +\frac{\kappa}{2} \norm{\vec{y}-\vec{x}}_2^2 \textrm{ for all }\vec{x},\vec{y} \in \mathbb{R}^d.
\end{equation}
It holds that
\begin{equation}
(\vec{y}-\vec{x})^T\left(\nabla f(\vec{y})-\nabla f(\vec{x})\right) \geq \kappa \norm{\vec{y}-\vec{x}}_2^2 \textrm{ for all }\vec{x},\vec{y} \in \mathbb{R}^d.
\end{equation}
\end{lemma}

\begin{lemma}\label{lem:wainwright910} Suppose that $f : \mathbb{R}^d \to \mathbb{R}$ is a twice differentiable, convex function that is locally $\kappa$-strongly convex around $\vec{x}$, in the sense that the lower bound \ref{eq:9.100a}  holds for all vectors $\vec{z}$ in the ball $\mathbb{B}_2 = \{\vec{z}: \norm{\vec{z}-\vec{x}}_2 \leq \rho\}$. It holds that
\begin{equation}
(\vec{y}-\vec{x})^T\left(\nabla f(\vec{y})-\nabla f(\vec{x})\right) \geq \rho\kappa \norm{\vec{y}-\vec{x}}_2 \textrm{ for all }\vec{y} \in \mathbb{R}^d \backslash \mathbb{B}_{2}.
\end{equation}
\end{lemma}
\begin{proof}
Let $\vec{y} \in \mathbb{R}^d \backslash \mathbb{B}_2$ and $\vec{x}_t = t(\vec{y}-\vec{x}) + \vec{x}$ and $g(t) = \nabla f(\vec{x}_t)$. The derivative is given by $g'(t) = \nabla ^2 f(\vec{x}_t) (\vec{y}-\vec{x})$. Let $0 \leq b \leq 1$ be such that $\norm{\vec{x}_b - \vec{x}}_2 = \rho$ and observe that $\rho = b\norm{\vec{y}-\vec{x}}_2$.
From fundamental theorem of calculus we get
\begin{align*}
(\vec{y} - \vec{x})^T(\nabla f(\vec{y}) - \nabla f(\vec{x})) &= (\vec{y} - \vec{x})^T(g(1) - g(0))\\
&=\int_0^{1} (\vec{y}-\vec{x})^T \nabla^2 f(\vec{x}_t) (\vec{y}-\vec{x}) dt\\&\geq
\int_0^{b} (\vec{y}-\vec{x})^T \nabla^2 f(\vec{x}_t) (\vec{y}-\vec{x}) dt \textrm{ since $f$ is convex}\\&\geq
\int_0^{b} \kappa\norm{\vec{y}-\vec{x}}_2^2 dt \textrm{ since $f$ is }\kappa-\textrm{strongly convex}\\&=
b\kappa\norm{\vec{y}-\vec{x}}_2^2 = \kappa \rho \norm{\vec{y}-\vec{x}}_2,
\end{align*}
and the claim follows.
\end{proof}

%% file: frobeniusv3.tex

\section{Statistics for Frobenius norm}\label{sec:frob}
\subsection{Graphical Lasso and Finite Population Likelihood}
The infinite population negative log-likelihood for a truncated Gaussian $\mathcal{N}\left(\vec{\mu}^*,\vec{\Sigma}^*\right)$ with variables $(\vec{\Theta},\vec{v})$ where $\vec{\Theta}$ captures $\vec{\Sigma}^{-1}$ and $\vec{v} = \vec{\Sigma}^{-1}\vec{\mu}$ is given by (see \cite{Das18} for calculations)
\begin{equation}\label{eq:population}
\overline{l}(\vec{\Theta},\vec{v}) := \mathbb{E}_{\vec{x} \sim\mathcal{N}\left(\vec{\mu}^*,\vec{\Sigma}^*,S\right)}\left[\frac{1}{2} \vec{x}^\top \vec{\Theta} \vec{x} - \vec{x}^\top\vec{v}\right] - \log \left(\int_S \exp(-\frac{1}{2} \vec{z}^\top \vec{\Theta}\vec{z}+\vec{z}^\top\vec{v})d\vec{z}\right).
\end{equation}

Moreover, the gradient of the function above $\overline{l}(\vec{\Theta},\vec{v})$ is given by
\begin{equation}\label{eq:derivativepopulation}
\nabla\overline{l}(\vec{\Theta},\vec{v}) := -\mathbb{E}_{\vec{x} \sim\mathcal{N}\left(\vec{\mu}^*,\vec{\Sigma}^*,S\right)}\left[\left(
\begin{array}{c}
\textrm{vec}(-\frac{1}{2}\vec{x}\vec{x}^\top)\\
\vec{x}
\end{array}
\right)\right] + \mathbb{E}_{\vec{z} \sim\mathcal{N}\left(\vec{\Theta}^{-1}\vec{v},\vec{\Theta}^{-1},S\right)}\left[\left(
\begin{array}{c}
\textrm{vec}(-\frac{1}{2}\vec{z}\vec{z}^\top)\\
\vec{z}
\end{array}
\right)\right]
\end{equation}
and its Hessian is
\begin{equation}\label{eq:derivativepopulation}
\nabla^2\overline{l}(\vec{\Theta},\vec{v}) := \textrm{Cov}_{\vec{z} \sim\mathcal{N}\left(\vec{\Theta}^{-1}\vec{v},\vec{\Theta}^{-1},S\right)}\left[\left(
\begin{array}{c}
\textrm{vec}(-\frac{1}{2}\vec{z}\vec{z}^\top)\\
\vec{z}
\end{array}
\right),
\left(
\begin{array}{c}
\textrm{vec}(-\frac{1}{2}\vec{z}\vec{z}^\top)\\
\vec{z}
\end{array}
\right)\right].
\end{equation}

We define the following score objective with parameter $\lambda > 0$ to be chosen later
\begin{equation}\label{eq:newfunction}
L_n(\vec{\Theta},\vec{v}):= l_n(\vec{\Theta},\vec{v}) + \lambda \norm{\textrm{vec}(\vec{\Theta})}_{1,\textrm{off}},
\end{equation}
where
\begin{equation}\label{eq:finitequadratic}
l_n(\vec{\Theta},\vec{v}) =  \frac{1}{n}\sum_{i=1}^n \frac{1}{2} \vec{x}_i^\top \vec{\Theta} \vec{x}_i - \frac{1}{n} \sum_{i=1}^n \vec{x}_i^\top \vec{v} -\log \left(\int_S \exp(-\frac{1}{2} \vec{z}^\top \vec{\Theta}\vec{z}+\vec{z}^\top\vec{v})d\vec{z}\right),
\end{equation}
given i.i.d samples $\vec{x}_1,...,\vec{x}_n$ from the true truncated distribution i.e., it is the finite population negative log-likelihood. We define $(\hat{\vec{\Theta}}, \hat{\vec{v}})$ to be a minimizer of $L_n$; this is the graphical Lasso estimator.

\subsection{Useful Lemmas}
In this section, we collect some results from previous work that will be crucial for us. The first establishes strong convexity
of the infinite population negative log-likelihood in a neighborhood around the true parameters.
\begin{lemma}[Lemma 4 and Lemma 7 in \cite{Das18}]\label{lem:strongdas} Let $H$ be the Hessian of the negative log-likelihood
function $\overline{l}(\vec{\Theta},\vec{v})$, with the presence of arbitrary truncation $S$ of measure $\alpha$ in the true truncated distribution. Assume that
\begin{enumerate}
\item $\norm{I - \vec{\Sigma}^{* 1/2}\vec{\Sigma}^{-1}\vec{\Sigma}^{* 1/2}}_F \leq B$.
\item $1/B \leq \norm{\vec{\Sigma}^{* -1/2}\vec{\Sigma}\vec{\Sigma}^{* -1/2}}_2 \leq B$.
\item $\norm{\vec{\Sigma}^{-1} \vec{\Sigma}^{* 1/2} (\vec{\mu}^* - \vec{\mu})}_2 \leq B$.
\end{enumerate}
It holds that there exists a constant $C$ so that
\[H \succeq C\left(\frac{\alpha}{12}\right)^{120B^5} \lambda_{\min}(\vec{\Sigma}^*) \vec{I}.\]
$B$ can be chosen to be $O\left(\frac{\log \left(1/\alpha\right)}{\alpha^2}\right)$.
\end{lemma}

We will also require the following concentration inequalities for the finite sample covariance matrix.
\begin{lemma}[Lemma 6.26 in \cite{Wbook}]\label{lem:wainwright626} Let $\vec{x}_1,...,\vec{x}_n$ be an i.i.d. sequence of $d$-dimensional zero-mean random vectors with covariance matrix $\vec{\Sigma}$, and suppose that each component $x_{ij}$ is a sub-Gaussian with parameter at most $\sigma$. If $n > \log d$, then for any $\delta>0$ we have
\begin{equation}
\Pr\left[\norm{\textrm{vec}(\vec{\Sigma} - \bar{\vec{\Sigma}})}_{\infty} \geq t \sigma^2\right] \leq 8 e^{-\frac{n}{16}\min(t,t^2)+ 2\log d} \textrm{ for all }t>0,
\end{equation}
where $\bar{\vec{\Sigma}}$ is the empirical covariance.
\end{lemma}

\begin{lemma}[Theorem 6.5 in \cite{Wbook}]\label{lem:covest}
There are universal constants $c_1, c_2, c_3$ such that for any row-wise $\sigma$-sub-Gaussian random matrix $\vec{X} \in \R^{n \times d}$, the sample covariance matrix $\hat{\vec{\Sigma}} = \frac1n \sum_{i=1}^n \vec{x}_i \vec{x}_i ^\top$ satisfies:
$$\Pr\left[\frac{\|\hat{\vec{\Sigma}}-\vec{\Sigma}\|_2}{\sigma^2} \geq c_1 \left\{\sqrt{\frac{d}{n}}+ \frac{d}{n}\right\} + \delta\right]\leq c_2 e^{-c_3 n \min(\delta,\delta^2)}$$
where $\vec{\Sigma}$ is the covariance matrix $\E[\vec{X}\vec{X}^\top]$.
\end{lemma}
Finally, we will use the following two lemmas that relate parameters of the truncated and untruncated Gaussian distribution.
\begin{lemma}\label{lem:covest2}
Suppose $\vec{x} \sim N(\vec{\mu},\vec{\Sigma})$ is a random vector in $\R^d$, and $F: \R^d \to \{0,1\}$ is a random function such that $\E_{\vec{x}}[\Pr[F(\vec{x})=1]] =\alpha$. Let $\vec{\mu}_F = \E[\vec{x} \cdot F(\vec{x})]$ and $\vec{\Sigma}_F = \E[\vec{x}\vec{x}^\top \cdot F(\vec{x})]$, and denote by $\hat{\vec{\mu}}_F$ and $\hat{\vec{\Sigma}}_F$ their respective empirical counterparts using $n = \tilde{O}(d\eps^{-2} \log \alpha^{-1} \log^2 \delta^{-1})$. Then, with probability $1-\delta$:
$$\|\vec{\Sigma}^{-1/2}(\hat{\vec{\mu}}_F-\vec{\mu}_F)\|_2\leq \eps \quad \text{and}\quad (1-\eps)\vec{\Sigma}_F \preceq \hat{\vec{\Sigma}}_F \preceq (1+\eps)\vec{\Sigma}_F.$$
\end{lemma}
\begin{proof}
The proof is exactly that of Lemma 5 in \cite{Das18}. The only difference is that $F$ is a random function here whereas $F$ is deterministic (indicator function of a subset) in \cite{Das18}. However, the proof remains unchanged.
\end{proof}
\begin{lemma}\label{lem:covest2}
Suppose $\vec{x} \sim N(\vec{\mu},\vec{\Sigma})$, and define $\vec{\mu}_F$ and $\vec{\Sigma}_F$ as in Lemma \ref{lem:covest2}. Then:
$$\|\vec{\mu}_F - \vec{\mu}\|_{\vec{\Sigma}} \leq O(\sqrt{\log \alpha^{-1}}) \quad \text{and} \quad
O(\alpha^{-2}) \vec{\Sigma} \succeq \vec{\Sigma}_F \succeq \Omega(\alpha^2) \vec{\Sigma}.$$
\end{lemma}

\subsection{Error Analysis}
From Lemma \ref{lem:strongdas} (one of the main Lemmas of \cite{Das18}), we know that $\overline{l}(\vec{v},\vec{\Theta})$ is strongly convex in some neighborhood $U \subseteq \mathcal{S}_{d\times d}\times \mathbb{R}^d$ of the true parameters. We can conclude that $L_n(\vec{\Theta},\vec{v})$ is also strongly convex in the same neighborhood because the term $\lambda \norm{\textrm{vec}(\vec{\Theta})}_{1,\textrm{off}}$ is a convex function and
\begin{equation}\label{eq:same}
\nabla^2 l_n(\vec{\Theta},\vec{v}) = \nabla^2 \overline{l}(\vec{\Theta},\vec{v}) \textrm{ i.e., they have same strong-convexity properties.}
\end{equation}
The following lemma indicates that the minimizer $\tilde{\vec{\Theta}},\tilde{\vec{v}}$ of function $L_n$ does not put too much weight on the coordinates $ij$ of $\tilde{\vec{\Theta}}$ for which $\vec{\Theta}^*_{ij} =0$, where $(\vec{\Theta}^*,\vec{v}^*)$ denote the true parameters.
\begin{lemma}[Lasso guarantee]\label{lem:main}
Let $(\tilde{\vec{\Theta}}, \tilde{\vec{v}})$ be the minimum of $L_n$ and $(\vec{\Theta}^*,\vec{v}^*)$ be the true parameters. Assume that $\lambda \geq 2\norm{\nabla_{\vec{\Theta}} l_n(\vec{\Theta}^*,\vec{v}^*)}_{\infty,\textrm{off}}$ and $\vec{\Delta} = \tilde{\vec{\Theta}} - \vec{\Theta}^*, \vec{\delta} = \tilde{\vec{v}} - \vec{v}^*$ then it holds
\[
\frac{1}{3}\norm{\textrm{vec}(\vec{\Delta}_{\overline{T}})}_1 - \frac{1}{3}\norm{\vec{\delta}}_1 \leq \norm{\textrm{vec}(\vec{\Delta}_{T})}_1,
\]
where $T$ denotes the support of $\vec{\Theta}^*$ and $\overline{T}$ denotes the complement. Moreover, we may assume that $\tilde{\vec{\Theta}}$ is symmetric.
\end{lemma}

From Lemma \ref{lem:main} and Cauchy-Schwarz inequality we conclude that
\begin{equation}\label{eq:csinequality}
\norm{\textrm{vec}(\vec{\Delta})}_1 + \norm{\delta}_1 \leq \norm{\textrm{vec}(\vec{\Delta}_T)}_1 + 3\norm{\textrm{vec}(\vec{\Delta}_T)}_1 + 2\norm{\delta}_1 \leq 4\sqrt{\textrm{nz}(\vec{\Theta}^*)+d}(\norm{\vec{\Delta}}_F+\norm{\delta}_2),
\end{equation}
We can now prove using Lemma \ref{lem:main} that for an appropriate choice of $\lambda$, the minimizer $(\tilde{\vec{\Theta}},\tilde{\vec{v}})$ of $L_n$ will be close to the true parameters $(\vec{\Theta}^*,\vec{v}^*)$.
\begin{lemma}[$(\tilde{\vec{\Theta}},\tilde{\vec{v}})$ are close to the true parameters]\label{lem:mainbig}
Let $L_n$ be $\kappa$-strong convex in a neighborhood of the true parameters. By choosing $\lambda$ to be $ O\left(\frac{\kappa \cdot \epsilon}{\sqrt{\textrm{nz}(\vec{\Theta}^*)}}\right)$ and moreover $\lambda \geq 2 \norm{\nabla l_n (\vec{\Theta}^*,\vec{v}^*)}_{\infty}$ then $\norm{\tilde{\vec{\Theta}} - \vec{\Theta}^*}_F + \norm{\tilde{\vec{v}} - \vec{v}^*}_2\leq \epsilon$.
\end{lemma}

We finish this section with a concentration lemma about how close the empirical mean and covariance is from the truncated mean and covariance in terms of $\ell_{\infty}$ and max norm respectively.
\begin{lemma}[Concentration of gradient] \label{lem:concentration1}
Assume that $n$ is $\Omega\left(\frac{\log d \log(1/\delta)}{t^2}\right)$ It holds that $$\Pr\left[\norm{\nabla l_n (\vec{\Theta}^*,\vec{v}^*)}_{\infty} \geq \frac{t}{2}\right] \leq \delta.$$
\end{lemma}

\subsection{Proof of Theorem \ref{thm:main1}}
We choose $\lambda$ to be $\tilde{O}\left(\frac{\epsilon}{12\sqrt{\textrm{nz}(\vec{\Theta}^*)+d}}\right)$ and consider the estimator $(\tilde{\vec{\Theta}},\tilde{\vec{v}}) := \arg\min_{\vec{\Theta},\vec{v}} L_n(\vec{\Theta},\vec{v}).$ 
\ignore{Notice that we need the oracle access to the truncation set $S$ so that we can compute the term $\mathbb{E}_{\vec{z} \sim\mathcal{N}\left(\vec{\Theta}^{-1}\vec{v},\vec{\Theta}^{-1},S\right)}\left[\left(
\begin{array}{c}
\textrm{vec}(-\frac{1}{2}\vec{z}\vec{z}^\top)\\
\vec{z}
\end{array}
\right)\right]$ which \textit{does not involve} the true parameters $(\vec{\Theta}^*,\vec{v}^*),$ (so we can approximate to arbitrary accuracy).  }
We will prove that $(\tilde{\vec{\Theta}},\tilde{\vec{v}})$ satisfies the statement of Theorem \ref{thm:main1}.

From Lemma \ref{lem:concentration1} we conclude that if $n$ is $\tilde{O}\left(\frac{(\textrm{nz}(\vec{\Theta}^*)+d)\log(1/\delta)}{\epsilon^2}\right)$ we get that $\lambda \geq 2 \norm{\nabla l_n (\vec{\Theta}^*,\vec{v}^*)}_{\infty}$ with probability $1-\delta$. Therefore the assumptions of Lemma \ref{lem:mainbig} hold and is guaranteed that the minimizer $(\tilde{\vec{\Theta}},\tilde{\vec{v}})$ of $L_n$ satisfies
    \begin{equation}
    \norm{\tilde{\vec{\Theta}}-\vec{\Theta}^*}_F \leq \epsilon \textrm{ and }\norm{\tilde{\vec{v}}-\vec{v}^*}_2 \leq \epsilon.
    \end{equation}

\subsection{Proof of Lemma \ref{lem:main}}
\begin{proof}
	It holds that
	\begin{align}\label{eq:inequalityzero}
		\norm{\textrm{vec}(\hat{\vec{\Theta}})}_{1,\textrm{off}} - \norm{\textrm{vec}(\vec{\Theta}^*)}_{1,\textrm{off}} &= \norm{\textrm{vec}(\vec{\Theta}^*) + \textrm{vec}(\vec{\Delta}_T)}_{1,\textrm{off}} + \norm{\textrm{vec}(\vec{\Delta}_{\overline{T}})}_{1,\textrm{off}}- \norm{\textrm{vec}(\vec{\Theta}^*)}_{1,\textrm{off}} \\&\geq \norm{\textrm{vec}(\vec{\Delta}_{\overline{T}})}_{1,\textrm{off}} - \norm{\textrm{vec}(\vec{\Delta}_{T})}_{1,\textrm{off}}
	\end{align}
	Observe that
	\begin{equation}\label{eq:addfirst}
	L_n(\vec{\Theta}^*, \vec{v}^*) - L_n(\hat{\vec{\Theta}}, \hat{\vec{v}})\geq 0.
	\end{equation}
	Moreover by convexity of $l_n$ we get that 
	\begin{align*}
		l_n(\hat{\vec{\Theta}}, \hat{\vec{v}}) - l_n(\vec{\Theta}^*, \vec{v}^*) &\geq (\vec{\Delta}^\top \ \vec{\delta}^\top) \nabla l_{n}(\vec{\Theta}^*, \vec{v}^*) \\
		& \geq - \left(\norm{\textrm{vec}(\vec{\Delta})}_{1,\textrm{off}}+ \norm{\vec{\delta}}_1\right) \cdot \norm{\nabla l_{n}(\vec{\Theta}^*, \vec{v}^*)}_{\infty,\textrm{off}},
	\end{align*}
	where the last inequality comes from Holder's inequality.
	Assuming that $\lambda \geq 2 \norm{\nabla l_n(\vec{\Theta}^*, \vec{v}^*)}_{\infty,\textrm{off}}$
	\begin{equation}\label{eq:second}
	l_n(\hat{\vec{\Theta}}, \hat{\vec{v}}) - l_n(\vec{\Theta}^*, \vec{v}^*) \geq - \frac{\lambda}{2}(\norm{\textrm{vec}(\vec{\Delta}_T)}_{1,\textrm{off}}+\norm{\textrm{vec}(\vec{\Delta}_{\overline{T}})}_{1,\textrm{off}} + \norm{\vec{\delta}}_1)
	\end{equation}
	We multiply (\ref{eq:inequalityzero}) by $\lambda$ and add it with (\ref{eq:addfirst}) and (\ref{eq:second}). It follows that
	\begin{equation}
	0 \geq \frac{\lambda}{2} \norm{\textrm{vec}(\vec{\Delta}_{\overline{T}})}_{1,\textrm{off}} - \frac{3\lambda}{2}\norm{\textrm{vec}(\vec{\Delta}_{T})}_{1,\textrm{off}} - \frac{\lambda}{2} \norm{\vec{\delta}}_1
	\end{equation}
	Therefore $\norm{\textrm{vec}(\vec{\Delta}_T)}_1 \geq \norm{\textrm{vec}(\vec{\Delta}_T)}_{1,\textrm{off}} \geq \frac{1}{3}\norm{\textrm{vec}(\vec{\Delta}_{\overline{T}})}_{1,\textrm{off}} - \frac{1}{3} \norm{\vec{\delta}}_1 = \frac{1}{3}\norm{\textrm{vec}(\vec{\Delta}_{\overline{T}})}_1 - \frac{1}{3} \norm{\vec{\delta}}_1$
	and the claim follows. To show that $\hat{\vec{\Theta}}$ is symmetric observe that if $(\vec{X},\hat{\vec{v}})$ is a minimum of $L_n$ by symmetry, so is $(\vec{X}^\top,\hat{\vec{v}})$. But $L\left(\frac{\vec{X}+\vec{X}^\top}{2},\hat{\vec{v}}\right) \leq \frac{1}{2} (L(\vec{X},\hat{\vec{v}})+L(\vec{X}^\top,\hat{\vec{v}}))$ by the triangle inequality for $\ell_1$ and the claim follows.
\end{proof}

\subsection{Proof of Lemma \ref{lem:mainbig}}
\begin{proof} We set $\vec{\Delta} = \hat{\vec{\Theta}} - \vec{\Theta}^*$, $\vec{\delta} = \hat{\vec{v}} - \vec{v}^*$. By the optimality of $(\hat{\vec{\Theta}},\hat{\vec{v}})$
	\begin{equation}
	(\textrm{vec}(\vec{\Delta})^\top \  \vec{\delta}^\top) (\nabla l_n(\hat{\vec{\Theta}},\hat{\vec{v}}) + \lambda \textrm{vec}(\vec{Z}))=0,
	\end{equation}
	where $\textrm{vec}(\vec{Z})$ is a subgradient for $\norm{\vec{\Theta}}_{1,\textrm{off}}$ computed at $\hat{\vec{\Theta}}$. Hence by Holder's inequality and taking absolute value for the subgradient part we get
	\begin{align}
		(\textrm{vec}(\vec{\Delta})^\top \  \vec{\delta}^\top) (\nabla l_n(\hat{\vec{\Theta}},\hat{\vec{v}}) - \nabla l_n(\vec{\Theta}^*,\vec{v}^*)) &\leq  \left(\norm{\textrm{vec}(\vec{\Delta})}_1 + \norm{\vec{\delta}}_1\right)\norm{\nabla l_n (\vec{\Theta}^*,\vec{v}^*)}_{\infty}-\lambda \textrm{vec}(\vec{\Delta})^\top(\textrm{vec}(\vec{Z})) \\&\label{eq:one}\leq  \lambda |\textrm{vec}(\vec{\Delta})^\top(\textrm{vec}(\vec{Z}))|+\frac{\lambda}{2} (\norm{\textrm{vec}(\vec{\Delta})}_1+\norm{\vec{\delta}}_1) \\&\leq \frac{3\lambda}{2} (\norm{\textrm{vec}(\vec{\Delta})}_1+ \norm{\vec{\delta}}_1).
	\end{align}
	Assume that $\norm{\vec{\Delta}}_F^2 + \norm{\vec{\delta}}_2^2 > r'^2$ where the ball $B\left(\begin{array}{c}\textrm{vec}(\vec{\Theta}^*) \\ \vec{v}^*\end{array}, r'\right)$ is a subset of $D_r$ as defined in \cite{Das18}. Observe that $r'$ is a function of $\alpha$ and $l_n$ is strongly convex with parameter $\kappa$ in the ball $B\left(\begin{array}{c}\textrm{vec}(\vec{\Theta}^*) \\ \vec{v}^*\end{array}, r'\right)$ (see Lemma \ref{lem:strongdas}). From Lemma \ref{lem:wainwright910} we get that:
	\begin{equation}\label{eq:two}
	(\textrm{vec}(\vec{\Delta})^\top \  \vec{\delta}^\top) (\nabla l_n(\hat{\vec{\Theta}},\hat{\vec{v}}) - \nabla l_n(\vec{\Theta}^*,\vec{v}^*)) \geq \kappa r' \sqrt{\norm{\vec{\Delta}}_F^2 + \norm{\vec{\delta}}^2_2} \geq \frac{\kappa r'}{2} (\norm{\vec{\Delta}}_F+\norm{\vec{\delta}}_2)
	\end{equation}
	Combining (\ref{eq:one}) and (\ref{eq:two}) along with (\ref{eq:csinequality}) we get that \[\frac{3\lambda}{2} \left(\norm{\textrm{vec}(\vec{\Delta})}_1 +\norm{\vec{\delta}}\right)\geq \frac{\kappa r'}{2} \left(\norm{\vec{\Delta}}_F+\norm{\vec{\delta}}_2\right) \geq \frac{\kappa r'}{8\sqrt{\textrm{nz}(\vec{\Theta}^*)+d}}(\norm{\textrm{vec}(\vec{\Delta})}_1+\norm{\vec{\delta}}_1).\]
	Therefore, if we choose $\lambda < \frac{\kappa r'}{12\sqrt{\textrm{nz}(\vec{\Theta}^*)+d}}$, we conclude that $\sqrt{\norm{\vec{\Delta}}_F^2 + \norm{\vec{\delta}}_2^2} \leq r'$.
	
	Now, by strong convexity of $l_n$ in that ball and Lemma \ref{lem:wainwright99} it follows that
	\begin{equation}\label{eq:finishproof}
	(\textrm{vec}(\vec{\Delta})^\top \  \vec{\delta}^\top)(\nabla l_n(\hat{\vec{\Theta}},\hat{\vec{v}}) - \nabla l_n(\vec{\Theta}^*,\vec{v}^*)) \geq \kappa \left(\norm{\vec{\Delta}}_F^2 + \norm{\vec{\delta}}^2_2\right) \geq \frac{\kappa}{2} \left(\norm{\vec{\Delta}}_F + \norm{\vec{\delta}}_2\right)^2.
	\end{equation}
	Hence again by combining the above with Equations (\ref{eq:one}) and (\ref{eq:csinequality}) we get that \[6\lambda \sqrt{\textrm{nz}(\vec{\Theta}^*)+d}(\norm{\vec{\Delta}}_F+\norm{\vec{\delta}}_2) \geq \frac{\kappa}{2} \left(\norm{\vec{\Delta}}_F + \norm{\vec{\delta}}_2\right)^2.\]
	Thus we choose $\lambda = \min\left(\frac{\kappa r'}{12\sqrt{\textrm{nz}(\vec{\Theta}^*)+d}}, \frac{\kappa \epsilon}{12\sqrt{\textrm{nz}(\vec{\Theta}^*)+d}} \right)$ and we conclude that $\norm{\vec{\Delta}}_F + \norm{\vec{\delta}}_2 \leq \epsilon$ and the proof is complete.
\end{proof}

\subsection{Proof of Lemma \ref{lem:concentration1}}
\begin{proof}
	Observe that
	\begin{equation}
	\begin{array}{c}
	\nabla_{\vec{\Theta}}l_n (\vec{\Theta}^*,\vec{v}^*) = \textrm{vec}\left(\frac{1}{2}\left(\bar{\vec{\Sigma}}_S+\bar{\vec{\mu}}_S\bar{\vec{\mu}}_S^\top - \vec{\Sigma}_S - \vec{\mu}_S\vec{\mu}_S^\top\right)\right)\\
	\nabla_{\vec{v}}l_n (\vec{\Theta}^*,\vec{v}^*) = \textrm{vec}\left(\vec{\mu}_S-\bar{\vec{\mu}}_S\right)
	\end{array}
	\end{equation}
	where $\bar{\vec{\mu}}_S,\bar{\vec{\Sigma}}_S$ are the empirical mean, covariance from $n$ samples from the true truncated distribution and $\vec{\mu}_S, \vec{\Sigma}_S$ are the mean and covariance matrix of the true truncated Gaussian.
	
	By an easy exact argument to Lemma 5 in \cite{Das18} and using Hoeffding’s inequality, we get that with $n$ at least $\Omega\left(\frac{\log (nd/\alpha\delta) \log (1/\delta)}{t^2}\right)$ it holds that 
	\[
	\Pr\left[\norm{\bar{\vec{\mu}}_S - \vec{\mu}_S}_{\infty} > \frac{t}{24}\right] \leq \frac{\delta}{2}.
	\]
	It is also clear that if $\norm{\bar{\vec{\mu}}_S - \vec{\mu}_S}_{\infty} \leq C$ and $C \leq 1$ then $\norm{\textrm{vec}(\bar{\vec{\mu}}_S\bar{\vec{\mu}}_S^\top - \vec{\mu}_S\vec{\mu}_S^\top)}_{\infty} \leq C (\norm{\bar{\vec{\mu}}_S}_{\infty}+ \norm{\vec{\mu}_S}_{\infty}) \leq 3C \max (1 , \norm{\vec{\mu}_S}_{\infty})$.
	Moreover, by a union bound argument one can show Lemma \ref{lem:wainwright626}, thus if
	$n$ at least $\Omega\left(\frac{\log d \log (1/\delta)}{t^2}\right)$ it holds that
	\[
	\Pr\left[\norm{\bar{\vec{\Sigma}}_S - \vec{\Sigma}_S}_{\infty} > \frac{t}{4}\right] \leq \frac{\delta}{2}.
	\]
	By adding the error probabilities and triangle inequality the claim follows.
\end{proof}

%% file: linreg.tex
\section{Sparse Linear Regression}\label{sec:linreg}

Recall the model described in the Introduction for the linear regression problem. The probability of obtaining a sample $(\vec{x},y) \in \R^d \times \R$ is:
$$\frac{\exp\left(-\frac12 (y-\vec{x}^\top {\vec{\Omega}^*})^2\right)}{\int  \exp\left(-\frac12 (z-\vec{x}^\top {\vec{\Omega}^*})^2\right) S(z) dz}$$
The infinite population negative log-likelihood function with $n$ samples is then:
\begin{equation}\label{eqn:popl}
\bar{l}(\vec{\Omega}) = \frac1n \sum_{i=1}^n \E_{y \sim \mathcal{N}(\vec{x}^{(i)}{\vec{\Omega}^*}, 1, S)}\left[\frac12 y^2 - y\cdot  \vec{x}^{(i)}\vec{\Omega} - \log \int \exp\left(-\frac12 z^2 + z \cdot  \vec{x}^{(i)}\vec{\Omega}\right) dz\right]
\end{equation}
where each $\vec{x}^{(i)}$ is viewed as a row vector.

As in the last section, we instead work with the finite sample negative log-likelihood, which is based on $n$ samples $(\vec{x}^{(1)}, y^{(i)}),\dots, (\vec{x}^{(n)},y^{(i)})$ with each $y^{(i)}$ being drawn from the distribution $\mathcal{N}(\vec{x}^{(i)}{\vec{\Omega}^*}, 1, S)$:
$$l_n(\vec{\Omega}) = \frac1n \sum_{i=1}^n \left(\frac{1}{2} {y^{(i)}}^2 - y^{(i)}  \vec{x}^{(i)}\vec{\Omega}  + \log \int \exp\left(-\frac{1}{2} z^2 +z  \vec{x}^{(i)}\vec{\Omega}\right) S(z) dz\right).$$

Note that $l_n$ is a random variable. We add a regularizer to  the sample negative log-likelihood to obtain the {\em truncated Lasso estimator}:
\begin{equation}\label{eqn:hat}
\hat{\vec{\Omega}} \in \arg \min_{\vec{\Omega} \in \R^d} \left\{l_n(\vec{\Omega}) + \lambda \|\vec{\Omega}\|_1\right\}.
\end{equation}
In the following, let $\vec{X}$ denote the $n$-by-$d$ design matrix whose $i$'th row corresponds to the $i$'th sample $\vec{x}^{(i)}$. Also,
 we let $\vec{x}_j \in \R^n$ denote the $j$'th column of $\vec{X}$.

\subsection{Assumptions} We now formally state the assumptions under which our result holds. For vectors $\vec{\Omega}$ and $\vec{x}$, let
$\alpha(\vec{\Omega}, \vec{x}) \defeq \E_{y \sim \mathcal{N}(\vec{\Omega}^\top \vec{x}, 1)}[S(y)].$
Also, in the following, let $K\subseteq [d]$ denote the support of $\vec{\Omega}^*$, and let $k = |K|$.

Our first assumption states that for every observed $\vec{x}^{(i)}$, there is a significant probability that the corresponding response variable $y^{(i)}$ is not truncated.
\begin{assumption}[Survival Probability]\label{ass:sur}
There exists a constant $\alpha>0$ such that for every $i \in [n]$, $\alpha(\vec{\Omega}^*, \vec{x}^{(i)}) \geq \alpha$.
\end{assumption}

Our second assumption is quite mild. It ensures that the model is identifiable when the support set $S$ is known in advance.
\begin{assumption}[Minimum Eigenvalue]\label{ass:id}
There exists a constant $\sigma_{\min}  > 0$ such that
$$\frac1n \vec{X}_K^\top \vec{X}_K \succeq \sigma_{\min}  \cdot \vec{I}.$$
\end{assumption}

Our third assumption ensures that the covariates corresponding to the support set are sufficiently prominent. More precisely,
the mutual incoherence assumption below requires that if $j \notin K$, then $\vec{x}_j$ is approximately orthogonal to the
span of the submatrix $\vec{X}_K$ corresponding to the covariates in $K$.
\begin{assumption}[Mutual incoherence]\label{ass:inc}
There exists a constant $\beta \in (0,1)$ such that:
$$\max_{j \notin K} \|\vec{x}_j^\top \vec{X}_K (\vec{X}_K^\top \vec{X}_K)^{-1}\|_1 \leq \beta.$$
\end{assumption}
Mutual incoherence is known to hold, for example, with high probability when $\vec{x}^{(1)}, \dots, \vec{x}^{(n)}$ are drawn i.i.d.~from $N(0,\vec{I}_{d \times d})$
as long as $n \geq \Omega\left(k \log d\right)$.

Our last assumption puts a bound on each entry of $\vec{X}$.
\begin{assumption}[Normalization]\label{ass:norm}
There exists  $C>0$ 
such that 
$\max_{i \in [n]} \|\vec{x}^{(i)}\|_\infty = \max_{j \in [d]}\|\vec{x}_j\|_\infty \leq C$.
\end{assumption}

\subsection{Support Recovery} We formally state the main theorem of this section.
\begin{theorem}\label{thm:regstat}
Consider a $k$-sparse linear regression model for which Assumptions \ref{ass:sur}, \ref{ass:id}, \ref{ass:inc}, and  \ref{ass:norm} are all satisfied.
Moreover, suppose that $\frac{C^2 k}{\alpha^5 \sigma_{\min}  (1-\beta)}$ is a sufficiently small constant.
Then, if
\[n \geq \tilde{\Omega}\left(\frac{C^4k^2\log d}{(1-\beta)^2\sigma_{\min} ^2 \alpha^9}\right)
\quad\quad
\text{and}
\quad\quad
\lambda  = \Theta\left(\frac{\alpha^4\sigma_{\min} }{Ck}\right),
\]
any solution $\hat{\vec{\Omega}}$ to the objective (\ref{eqn:hat}) satisfies the following properties with high probability.
\begin{itemize}
\item[(a)]
\textbf{Uniqueness}: There is a unique solution $\hat{\vec{\Omega}}$.
\item[(b)]
\textbf{No false inclusion}: $\mathrm{supp}(\hat{\vec{\Omega}}) \subseteq \mathrm{supp}(\vec{\Omega}^*)$.
\item[(c)]
\textbf{$\ell_\infty$-bounds}: The error $\hat{\vec{\Omega}}-\vec{\Omega}^*$ satisfies
\[
\|\hat{\vec{\Omega}}-\vec{\Omega}^*\|_\infty \leq O\left(\sqrt{\frac{\log(1/\alpha)}{\sigma_{\min} }} + \frac{\alpha^4}{C\sqrt{k}} \right).
\]
\end{itemize}
\end{theorem}
In other words, if the non-zero entries of $\vec{\Omega}^*$ are greater than a particular threshold $\tau$ (which is independent of $d$), then
the support of $\hat{\vec{\Omega}}$ exactly matches with the support of $\vec{\Omega}^*$.

In the untruncated setting, it is known (see Chapter 7 of \cite{Wbook}) that $\lambda$ can be made to scale as $\sim \frac{1}{\sqrt{n}}$, and the 
$\ell_\infty$ error is the sum of two terms, one proportional to $\lambda$ and the other to $\frac{1}{\sqrt{n}}$. Hence, by making $n$ large, the
$\ell_\infty$ error can be made arbitrarily small. In contrast, in our analysis, we cannot make $\lambda$ arbitrarily small; so, above, we fix it in terms of the
other problem parameters. 

The other notable aspect of Theorem \ref{thm:regstat} is the hypothesis that $\frac{C^2 k}{\alpha^5 \sigma_{\min}  (1-\beta)}$ is small, which is
also absent from the untruncated setting. The hypothesis can be satisfied if $C$ is mildly decreasing in $d$ (e.g., $\sim 1/\log(d)$), and $d$ is very large. 

\subsection{Useful Lemmas}
We record for later use two convenient lemmas from \cite{Das19}.
\begin{lemma}[Lemma 6 in \cite{Das19}]\label{lem:das2}
For a set $S \subseteq \R$ and vectors $\vec{\Omega}, \vec{\Omega}', \vec{x} \in \R^{d}$, :
$$\alpha(\vec{\Omega},\vec{x}) \geq \alpha(\vec{\Omega}',\vec{x})^2 \cdot \exp\left(-|(\vec{\Omega}-\vec{\Omega}')^\top \vec{x}|^2 - 2\right).$$
\end{lemma}

\begin{lemma}[Lemma 14 in \cite{Das19}]\label{lem:das1}
For $\vec{x}, \vec{w} \in \R^k$, if $z \sim N(\vec{w}^\top \vec{x}, 1, S)$, then:
$$\E[(z-\E[z])^2] \geq \frac{1}{12}\left(\int_S N(\vec{w}^\top \vec{x}, 1)(y) \cdot S(y) \cdot dy\right)^2.
$$
\end{lemma}

\input{linreg-proofs}

%% file: linreg-proofs.tex
\subsection{Proof of Theorem \ref{thm:regstat}}
Standard calculations (see, e.g., \cite{Das18, Das19}) show that the gradient and Hessian of the empirical log-likelihood can be written as:
\begin{equation}\label{eqn:reggrad}
\nabla l_n(\vec{\Omega}) = -\frac1n \sum_{i=1}^n  \left(\vec{x}^{(i)}y^{(i)} - \E_{z^{(i)} \sim \mathcal{N}( \vec{x}^{(i)}\vec{\Omega}, 1, S)}[\vec{x}^{(i)} z^{(i)}]\right)
\end{equation}
\begin{equation}\label{eqn:reghess}
H(\vec{\Omega}) \defeq \nabla^2 l_n(\vec{\Omega}) = \frac1n \sum_{i=1}^n \Cov_{z^{(i)} \sim \mathcal{N}( \vec{x}^{(i)}\vec{\Omega}, 1, S)}[\vec{x}^{(i)} z^{(i)}, \vec{x}^{(i)} z^{(i)}]
\end{equation}
From (\ref{eqn:reghess}), since $H(\vec{\Omega}) \succeq 0$, it is clear that $l_n$ is convex for arbitrary choices of $\vec{x}^{(1)}, \dots, \vec{x}^{(n)}$.
\ignore{
Also, note that the Hessian of $l_n$ and $\bar{\ell}$ (from (\ref{eqn:popl})) are identical because neither depend on the $y$ variable. Hence, the strong convexity property of $\bar{\ell}, l_n$ and $l_n+\lambda \|\cdot\|_1$ are the same.


\begin{proof}[Proof of Theorem \ref{thm:regstat}]
The proof follows the outline of Wainwright's proof in \cite{Wbook} (Chapter 7) for subset selection by Lasso, except that our likelihood involves the truncation set, which significantly complicates the analysis.
}
Let:
$$F(\vec{\Omega}) = l_n(\vec{\Omega}) + \lambda \|\vec{\Omega}\|_1.$$
Clearly, $F$ is also convex since $l_n$ is. Now, any optimum $\hat{\vec{\Omega}} \in \arg \min F(\vec{\Omega})$ has to satisfy the zero-subgradient condition:
\begin{equation}\label{eqn:zsub}
\nabla l_n(\hat{\vec{\Omega}}) + \lambda \hat{\vec{W}} = 0
\end{equation}
where $\hat{\vec{W}} \in \partial \|\hat{\vec{\Omega}}\|_1$ and $\nabla l_n$ is as in (\ref{eqn:reggrad}).

Recall that $K \subseteq [d]$ denotes the support of $\vec{\Omega}^\ast$ and $|K| = k$. Let
\begin{equation}\label{eqn:subprob}
\check{\vec{\Omega}} \in \arg \min_{\vec{\Omega} \in \R^{d-1}: \forall j \notin K, \Omega_j = 0} F(\vec{\Omega})
\end{equation}
The rest of the proof goes as follows. First, we establish strong convexity of the restricted likelihood function:
\[
l_n^{K}(\vec{\Omega}) = \frac1n \sum_{i=1}^n \left(\frac{1}{2} {y^{(i)}}^2 -  \vec{x}^{(i)}\vec{\Omega}_{K} y^{(i)} - \ln \int \exp\left(-\frac{1}{2} z^2+ \vec{x}^{(i)}_{K}\vec{\Omega}  z\right) S(z) dz\right),
\]
when $\vec{\Omega}$ lies in a neighborhood of $\vec{\Omega}^*$.
We show that our choice of $\lambda$ implies that $\check{\vec{\Omega}}$ falls in this neighborhood, and hence, there is a unique choice of $\check{\vec{\Omega}}$ in (\ref{eqn:subprob}).
Next, we use the {primal-dual witness method} to show that if strict dual feasibility holds, then $\hat{\vec{\Omega}}$ must in fact equal $\check{\vec{\Omega}}$.
 Then, we verify that strict dual feasibility holds under our choice of parameters and assumptions, establishing parts (a) and (b) of Theorem \ref{thm:regstat}. Finally, it remains to bound the $\ell_\infty$-distance between $\check{\vec{\Omega}}$ and $\vec{\Omega}^*$, proving part (c).

\ignore{
\begin{lemma}\label{lem:subunique}
If $n \geq {\Omega}(m^2 \cdot \mathrm{polylog}(mn/\alpha \delta))$, then with probability at least $1-\delta$, $\check{\vec{\Omega}}$ in (\ref{eqn:subprob}) is uniquely defined.
\end{lemma}
\begin{proof}
Define the restricted likelihood function $l_n^{T}: \R^{|T|} \to \R$:}

\paragraph{Local strong convexity:}
We first show that $l_n^K$ is strongly convex for $\vec{\Omega}$ within a ball centered at $\vec{\Omega}^*$. Below, $\vec{\Omega}^\ast_{K}$ is the restriction of $\vec{\Omega}^\ast$ to the coordinates in $K$.
\begin{lemma}\label{lem:locstrong}
There exists $\kappa\geq \Omega( \alpha^4\cdot \sigma_{\min} )>0$ such that $\nabla^2l_n^{K}(\vec{\Omega}) \succeq \kappa\cdot \vec{I}$ for all $\vec{\Omega}\in \R^{k}$ satisfying $\|\vec{\Omega}-\vec{\Omega}_{K}^*\|_2\leq \frac{1}{C\sqrt{k}}$.
\end{lemma}
\begin{proof}
	Computing the Hessian of $l_n^{K}$:
	\begin{align*}
		\nabla^2 l_n^{K}(\vec{\Omega}) &= \frac1n \sum_{i=1}^n \Cov_{z^{(i)} \sim \mathcal{N}(\vec{x}^{(i)}_K\vec{\Omega},1,S)}[\vec{x}^{(i)}_Kz^{(i)},\vec{x}^{(i)}_Kz^{(i)}]\\
		&= \frac1n \sum_{i=1}^n {\vec{x}_K^{(i)}}^\top {\vec{x}_K^{(i)}} \cdot \E_{z^{(i)} \sim \mathcal{N}(\vec{x}^{(i)}_K\vec{\Omega},1,S)}[(z^{(i)}-\E z^{(i)})^2]\\
		&\succeq  \frac{1}{12n} \sum_{i=1}^n {\vec{x}_K^{(i)}}^\top {\vec{x}_K^{(i)}} \cdot (\alpha(\vec{\Omega},\vec{x}^{(i)}))^2\\
		&\succeq \frac{1}{12e^2n} \sum_{i=1}^n {\vec{x}_K^{(i)}}^\top {\vec{x}_K^{(i)}} (\alpha(\vec{\Omega}^*, \vec{x}^{(i)}))^4 \cdot \exp\left(-2|\vec{x}^{(i)}_K(\vec{\Omega}-{\vec{\Omega}^*_{K}})|^2-4\right)
	\end{align*}
	where the third line\footnote{We abuse notation in the third line. When we say $\alpha(\vec{\Omega},\vec{x}^{(i)})$, we mean by $\vec{\Omega}$ the extension of the vector to $\R^d$ where the coordinates not in $K$ are set to zero.} follows from Lemma \ref{lem:das1} and the fourth line follows from Lemma \ref{lem:das2}. Recall that by Assumption \ref{ass:norm},  $\|\vec{x}_K^{(i)}\|_\infty \leq C$, and hence, $\|\vec{x}_K^{(i)}\|_2 \leq C\sqrt{k}$. Plugging into the above, and again using Assumptions \ref{ass:sur} and \ref{ass:id}, we obtain:
	\[\nabla^2 l_n^{K}(\vec{\Omega}) \succeq \frac{1}{12e^2} \cdot \alpha^4 \cdot e^{-6} \cdot \sigma_{\min}  \vec{I}.\]
\end{proof}

\paragraph{Bounding the $\ell_2$-distance between $\check{\vec{\Omega}}$ and $\vec{\Omega}^*$:}
Let $r = \frac{1}{C\sqrt{k}}$ be the radius of the ball $B_r$ around $\vec{\Omega}_{K}^*$ inside which $l_n^{K}$ is strongly convex (by Lemma \ref{lem:locstrong}). This means that $l_n^{K}(\vec{\Omega})+\lambda  \|\vec{\Omega}\|_1$ is also locally strongly convex in $B_r$. The next proposition shows that for an appropriate choice of $\lambda $, $\check{\vec{\Omega}}$, defined in (\ref{eqn:subprob}) as the minimizer of $l_n^{K}(\vec{\Omega})+\lambda  \|\vec{\Omega}\|_1$, is in the ball $B_r$.

\begin{lemma}\label{lem:lambda}
If $2\|\nabla l_n^{T}(\vec{\Omega}^*_{K})\|_\infty < \lambda  <\frac{2\kappa r}{3\sqrt{k}}$, where $\kappa$ and $r$ are as above, then $\check{\vec{\Omega}} \in B_r$. Moreover, our choice of $\lambda $ satisfies these conditions with high probability.
\end{lemma}
\begin{proof}
	The first part follows exactly the same steps as the proof of Lemma \ref{lem:mainbig}, and so we omit it. For the second part, note that:
	\[\nabla l_n^{K}(\vec{\Omega}^*_{K}) = -\frac1n \sum_{i=1}^n \left(\vec{x}^{(i)}_K y^{(i)} - \E_{z^{(i)} \sim \mathcal{N}(\vec{x}_{K}^{(i)}{\vec{\Omega}^*_{K}}, 1, S)}[\vec{x}_{K}^{(i)} z^{(i)}]\right).\]
Consider the $i$'th summand $\vec{x}^{(i)}_K\cdot \left(y^{(i)} - \E_{z^{(i)} \sim \mathcal{N}(\vec{x}_{K}^{(i)}{\vec{\Omega}^*_{K}}, 1, S)}[z^{(i)}]\right)$. For every $i$, each coordinate of the $i$'th summand has mean zero. Also, for every $i$, each coordinate of the $i$'th summand is bounded by $O(C \sqrt{\log(n/\alpha)})$ with high probability because:
	\begin{itemize}
	\item[(i)] $\|\vec{x}_K^{(i)}\|_\infty \leq C$ for all $i$ by Assumption \ref{ass:norm}.
\item[(ii)]	 $y^{(i)}-\vec{x}_K^{(i)}\vec{\Omega}_K^* $ is distributed as a standard normal truncated to a set of volume at least $\alpha$. The maximum of $n/\alpha$ standard normal variables is $O(\sqrt{\log(n/\alpha)})$ with high probability.
\item[(iii)] $\vec{x}_K^{(i)}\vec{\Omega}_K^*  - \E[\mathcal{N}(\vec{x}_{K}^{(i)}{\vec{\Omega}^*_{K}}, 1, S)]= \E[\mathcal{N}(\vec{x}_{K}^{(i)}{\vec{\Omega}^*_{K}}, 1)] - \E[\mathcal{N}(\vec{x}_{K}^{(i)}{\vec{\Omega}^*_{K}}, 1, S)]$ is bounded by $O(\sqrt{\log(1/\alpha)})$ by Lemma \ref{lem:covest2}.
\end{itemize}

By the Hoeffding bound, $\|\nabla \ell_{K}(\vec{\Omega}^*_{K})\|_\infty < t$ with high probability if $n \geq {\Omega}(C^2\log k\log(n/\alpha)/t^2)$. Plugging in $t = O(\kappa r/\sqrt{k})$ and the values of $\kappa$ and $r$ from Lemma \ref{lem:locstrong}, we see that the constraint is non-vacuous if $n\geq \tilde{\Omega}(C^2/(\kappa^2r^2/k)) = \tilde{\Omega}(C^4k^2/\alpha^8\sigma_{\min} ^2)$. The condition on $n$ is satisfied by the hypotheses of Theorem \ref{thm:regstat}.
\end{proof}
\noindent The above two lemmas imply that $\check{\vec{\Omega}}$ is uniquely defined. Now, the goal is to relate it to the structure of $\hat{\vec{\Omega}}$.
\paragraph{Unique global minimum under strict dual feasibility:}
We construct a vector $\check{\vec{W}}$ so that $(\check{\vec{\Omega}}, \check{\vec{W}})$ satisfy the zero subgradient condition (\ref{eqn:zsub}). Note that by definition, $\check{\vec{\Omega}}$ satisfies the restricted zero-subgradient condition on the coordinates in $K$:
$$\frac1n \sum_{i=1}^n \vec{x}_K^{(i)} y^{(i)} - \E_{z^{(i)} \sim \mathcal{N}(\vec{x}^{(i)}_K\check{\vec{\Omega}},1,S))}[\vec{x}_{K}^{(i)} z^{(i)}] - \lambda \check{\vec{W}}_{K} = 0$$
for some $\check{\vec{W}}_{K} \in \partial \|\check{\vec{\Omega}}_{K}\|_1$. We set $\check{\vec{W}}$ so that it restricts to $\check{\vec{W}}_{K}$ on $K$ and satisfies the zero-subgradient condition (\ref{eqn:zsub}) on all other coordinates as well.

\begin{lemma}\label{lem:uniquemin}
If $\|\check{\vec{W}}_{-K}\|_\infty < 1$, then $\check{\vec{\Omega}} = \hat{\vec{\Omega}}$ is the unique minimizer of $F$.
\end{lemma}
\begin{proof}
Note that $\langle \check{\vec{W}}, \check{\vec{\Omega}} \rangle = \|\check{\vec{\Omega}}\|_1$ because outside $K$, $\check{\vec{\Omega}}$ is zero, and $\check{\vec{W}}_{K} \in \partial \|\vec{\Omega}_{K}\|_1$. Then, since $F(\hat{\vec{\Omega}}) \leq F(\check{\vec{\Omega}})$:
\begin{align*}
\lambda \|\hat{\vec{\Omega}}\|_1 &\leq l_n(\check{\vec{\Omega}}) + \lambda \|\check{\vec{\Omega}}\|_1 - l_n(\hat{\vec{\Omega}})\\
&= l_n(\check{\vec{\Omega}}) - l_n(\hat{\vec{\Omega}}) + \langle \lambda \check{\vec{W}}, \check{\vec{\Omega}}\rangle \\
&= l_n(\check{\vec{\Omega}}) - l_n(\hat{\vec{\Omega}}) -  \langle \nabla l_n(\check{\vec{\Omega}}), \check{\vec{\Omega}}\rangle
\end{align*}
where we used the fact that $\check{\vec{\Omega}}$ and $\check{\vec{W}}$ satisfy the zero-subgradient condition (\ref{eqn:zsub}). Invoking the convexity of $l_n$:
\begin{align*}
\lambda \|\hat{\vec{\Omega}}\|_1&\leq - \langle \nabla \bar{l_n}(\check{\vec{\Omega}}), \hat{\vec{\Omega}}\rangle = \lambda \langle \check{\vec{W}}, \hat{\vec{\Omega}}\rangle
\end{align*}
Hence, if $|\check{\vec{W}}_{K}| < 1$ for any $s$, then $\hat{\vec{\Omega}}_{K} = 0$. The claim then follows since we have already established that $\check{\vec{\Omega}}$ is defined uniquely.
\end{proof}

\paragraph{Verifying strict dual feasibility:} We next confirm that the $\check{\vec{W}}$ vector constructed above satisfies the condition of Lemma \ref{lem:uniquemin}.
\begin{lemma}
With high probability, $\|\check{\vec{W}}_{-K}\|_\infty<1$.
\end{lemma}
\begin{proof}
Since $(\check{\vec{\Omega}}, \check{\vec{W}})$ satisfy the zero subgradient condition (\ref{eqn:zsub}), we can solve for $\check{\vec{W}}_{-K}$:
\begin{align}
\check{\vec{W}}_{-K}
&=\frac{1}{n\lambda }\sum_{i=1}^n {\vec{x}^{(i)}_{-K}}^\top \left(y^{(i)} - \E_{z^{(i)} \sim \mathcal{N}( \vec{x}^{(i)}\check{\vec{\Omega}}, 1, S)}z^{(i)}\right)\nonumber\\
&=\frac{1}{n\lambda } \sum_i {\vec{x}^{(i)}_{-K}}^\top \left({\vec{x}_{K}^{(i)}} (\vec{\Omega}^* - \check{\vec{\Omega}})+ {z^*}^{(i)} - \E \check{z}^{(i)}\right)\label{eqn:j-K}
\end{align}
where ${z^*}^{(i)} \sim N(0,1, S(\cdot + \vec{\Omega}^* \vec{x}^{(i)}))$ and ${\check{z}}^{(i)}\sim N(0,1, S(\cdot +  \vec{x}^{(i)}\check{\vec{\Omega}}))$. Similarly, we solve for $\check{\vec{W}}_{K}$:
\begin{align*}
\check{\vec{W}}_{K}
&=\frac{1}{n\lambda }\sum_{i=1}^n {\vec{x}^{(i)}_{K}}^\top \left(y^{(i)} - \E_{z^{(i)} \sim \mathcal{N}(\vec{x}^{(i)}\check{\vec{\Omega}}, 1, S)}z^{(i)}\right)\nonumber\\
&=\frac{1}{n\lambda } \sum_i {\vec{x}^{(i)}_{K}}^\top \left({\vec{x}_{K}^{(i)}} (\vec{\Omega}^*-\check{\vec{\Omega}}) + {z^*}^{(i)} - \E \check{z}^{(i)}\right)\label{eqn:j-K}
\end{align*}
Let $\vec{u}$ denote the random vector whose $i$'th coordinate is ${z^*}^{(i)} - \E \check{z}^{(i)}$; note that the components of $\vec{u}$ are independent.  We can rewrite the above as:
\begin{equation}\label{eqn:omegadiff}
\vec{\Omega}^*-\check{\vec{\Omega}}  = \lambda n (\vec{X}_K^T \vec{X}_K)^{-1} \check{\vec{W}}_K - (\vec{X}_{K}^\top \vec{X}_K)^{-1}\vec{X}_K^\top \vec{u}
\end{equation}
Substituting back into (\ref{eqn:j-K}), we get:
\begin{equation}\label{eqn:boundonw}
\vec{W}_{-K} = \vec{X}_{-K}^\top \vec{X}_K (\vec{X}_K^\top \vec{X}_K)^{-1} \check{\vec{W}}_K + \frac{1}{\lambda n} \vec{X}_{-K}^\top (\vec{I}-\vec{X}_K(\vec{X}_K^\top \vec{X}_K)^{-1} \vec{X}_K^\top) \vec{u}.
\end{equation}
We analyze the contribution of each of the two terms above separately.
\begin{claim}
$\|\vec{X}_{-K}^\top \vec{X}_K (\vec{X}_K^\top \vec{X}_K)^{-1} \check{\vec{W}}_K\|_\infty \leq \beta$.
\end{claim}
\begin{proof}
For any $j \notin K$,
\[|\vec{x}_j^\top \vec{X}_K (\vec{X}_K^\top \vec{X}_K)^{-1} \check{\vec{W}}_K| \leq \|\vec{x}_j^\top \vec{X}_K (\vec{X}_K^\top \vec{X}_K)^{-1}\|_1 \leq \beta\]
by Assumption \ref{ass:inc}.
\end{proof}
\begin{claim}\label{clm:randdual}
With high probability, $\|\frac{1}{\lambda n} \vec{X}_{-K}^\top (\vec{I}-\vec{X}_K(\vec{X}_K^\top \vec{X}_K)^{-1} \vec{X}_K^\top) \vec{u}\|_\infty \leq \frac{1-\beta}{2}.$
\end{claim}
\begin{proof}
For any $j \notin K$, let $\vec{v}_j^\top = \vec{x}_j^\top (\vec{I}-\vec{X}_K(\vec{X}_K^\top \vec{X}_K)^{-1} \vec{X}_K^\top)$. Note that $\vec{I}-\vec{X}_K(\vec{X}_K^\top \vec{X}_K)^{-1} \vec{X}_K^\top$ is an
orthogonal projection matrix, and hence, by Assumption \ref{ass:norm}, $\|\vec{v}_j\|_2 \leq \|\vec{x}_j\|_2 \leq C\sqrt{n}$ and $\|\vec{v}_j\|_1 \leq Cn$.

Write $\vec{u} = \vec{z}^* - \E[\check{\vec{z}}]$. By the triangle inequality, it suffices to bound: (i) $\frac{1}{\lambda n} |\vec{v}_j^\top (\vec{z}^* - \E[\vec{z}^*])|$ and (ii) $\frac{1}{\lambda n} |\vec{v}_j^\top (\E[\vec{z}^*] - \E[\check{\vec{z}}])|$.

We first bound (ii).
\[
\frac{1}{\lambda n} |\vec{v}_j^\top (\E[\vec{z}^*] - \E[\check{\vec{z}}])| \leq \frac{1}{\lambda n} \|\vec{v}_j\|_1 \|\E[\vec{z}^*] - \E[\check{\vec{z}}])\|_\infty \leq \frac{C}{\lambda } \cdot\max_{i \in [n]} (|\E[{z^*}^{(i)}]| + |\E[\check{z}^{(i)}]|)
\]
Since each ${z^*}^{(i)}$ is a standard normal truncated to a set of volume at least $\alpha$, Lemma \ref{lem:covest2} implies $\E[{z^*}^{(i)}] = O(\sqrt{\log(1/\alpha)})$.  Similarly, $\E[\check{z}^{(i)}] = O\left(\sqrt{\log(1/\alpha(\check{\vec{\Omega}}_K, \vec{x}_K^{(i)}))}\right)$. Invoking the fact that $\|\vec{\Omega}^* - \check{\vec{\Omega}}\|_2 \leq \frac{1}{C\sqrt{k}}$ and Lemma \ref{lem:das2}, we obtain that $\alpha(\check{\vec{\Omega}}, \vec{x}^{(i)}) = \Omega(\alpha^2)$. Hence, we get:
\[
\frac{1}{\lambda n} |\vec{v}_j^\top (\E[\vec{z}^*] - \E[\check{\vec{z}}])|\leq \frac{C}{\lambda } \cdot O(\sqrt{\log(1/\alpha)}) \leq O\left(\frac{C^2k}{\alpha^5 \sigma_{\min} }\right) < \frac{1-\beta}{4}.
\]
where we used the assumption in the Theorem.

Now, we turn to (i). Observe that $\vec{\zeta}^* = \vec{z}^* - \E[\vec{z}^*]$ is a zero-mean vector with independent components ${\zeta^*}^{(1)}, \dots, {\zeta^*}^{(n)}$. In order to bound $\vec{v}_j^\top \vec{\zeta}^* = \sum_{i \in [n]} {v}_j^{(i)} {\zeta^*}^{(i)}$, we use Bernstein's inequality. Fix an $i \in [n]$, and let $\zeta$ denote ${\zeta^*}^{(i)}$. For Bernstein's Lemma, we need bounds on $\E[{{\zeta}}^p]$ for $p>1$. It is easy to see that these quantities are maximized when $\zeta \sim \mathcal{N}(0,1,S_q)$ where $S_q = \{x : x^2 \geq q\}$ for some $q$ chosen such that $\mathcal{N}(0,1,S_q)=\alpha$. Routine calculations (Lemma 13 of \cite{Das19})  show that $\E[\zeta^2] \leq 2 + 2 \log(2/\alpha)$ and $\E[\zeta^p] \leq p! (2+2\log(2/\alpha))^p$. Applying Bernstein, we get that with probability $1-\exp(-t^2)$,  $\frac{1}{\lambda n} |\vec{v}_j^T \vec{\zeta}^*| \leq O\left(\frac{Ct\log(1/\alpha)}{\lambda \sqrt{n}}\right)$ for $t \leq O(\sqrt{n/\log(1/\alpha)})$.  Setting $t = \Omega(\sqrt{\log d})$, we get that with high probability, for all $j \notin K$:
\[
\frac{1}{\lambda n} |\vec{v}_j^T \vec{\zeta}^*| \leq O\left(\frac{C\sqrt{\log d}\log(1/\alpha)}{\lambda \sqrt{n}}\right) = O\left(\frac{C^2 k \sqrt{\log d}\log(1/\alpha)}{\alpha^4 \sigma_{\min}  \sqrt{n}}\right).
\]
If we take $n \geq \Omega((C^4k^2\log d)/((1-\beta)^2\sigma_{\min} ^2 \alpha^9))$, then the above is less than $(1-\beta)/4$. This proves the claim.
\end{proof}

\paragraph{Bounding the $\ell_\infty$-error:} The max-error can be bounded using (\ref{eqn:omegadiff}).
\[
\|\check{\vec{\Omega}}-\vec{\Omega}^*\|_\infty \leq \lambda  \opnorm*{\left(\frac{\vec{X}_K^\top \vec{X}_K}{n}\right)^{-1}}_\infty + \left\|\left(\frac{\vec{X}_K^\top \vec{X}_K}{n}\right)^{-1}\vec{X}_K^\top \frac{\vec{u}}{n}\right\|_\infty
\]
where for a matrix $A$ with $r$ rows, $\opnorm{A}_\infty = \max_{i \in [r]} \|A^{(i)}\|_1$ is the matrix $\ell_\infty$-norm.

The first term is deterministic and can be bounded as follows:
\[
\lambda  \opnorm*{\left(\frac{\vec{X}_K^\top \vec{X}_K}{n}\right)^{-1}}_\infty \leq \lambda  \sqrt{k}\cdot \lambda_{\max}\left(\left(\frac{\vec{X}_K^\top \vec{X}_K}{n}\right)^{-1}\right) \leq \frac{\lambda  \sqrt{k}}{\sigma_{\min} } = O\left(\frac{\alpha^4}{C\sqrt{k}}\right)
\]

To analyze the second term, define for $j \in K$, the vector $\vec{w}_j^\top = \vec{e}_j^\top \left(\frac{\vec{X}_K^\top \vec{X}_K}{n}\right)^{-1}\frac{\vec{X}_K^\top}{n}$. We need to bound $\max_{j\in K} |\vec{w}_j^\top \vec{u}|$.
Note that $\|\vec{w}_j\|_2^2 = 	\frac1n \vec{e}_j^\top \left(\frac{\vec{X}_K^\top \vec{X}_K}{n}\right)^{-1}\vec{e}_j \leq \frac{1}{n \sigma_{\min} }$.
Similar to the analysis in the proof of Claim \ref{clm:randdual}, we write $\vec{u}$ as $(\vec{z}^* - \E[\vec{z}^*]) + (\E[\vec{z}^*] - \E[\check{\vec{z}}])$. The quantity $\max_{j \in K}|\vec{w}_j^\top (\vec{z}^* - \E[\vec{z}^*])|$ can be
shown using Bernstein's inequality to be at most $O\left(\frac{\sqrt{\log k}\log(1/\alpha)}{\sqrt{\sigma_{\min}  n}}\right)$ with high probability. The other term can be bounded as:
\[
\max_{j \in K} |\vec{w}_j^\top (\E[\vec{z}^*] - \E[\check{\vec{z}}])| \leq \max_{j \in K} \|\vec{w}_j\|_1 O(\sqrt{\log(1/\alpha)}) \leq O\left(\sqrt{\frac{\log(1/\alpha)}{\sigma_{\min} }}\right)
\]
So, putting everything together:
\[
\|\check{\vec{\Omega}}-\vec{\Omega}^*\|_\infty \leq O\left(\sqrt{\frac{\log(1/\alpha)}{\sigma_{\min} }} + \frac{\alpha^4}{C\sqrt{k}} + \frac{\sqrt{\log k}\log(1/\alpha)}{\sqrt{\sigma_{\min}  n}} \right).
\]
The last term is negligible because of the lower bound on $n$.
\ignore{
 The $j$'th component of the second term in (\ref{eqn:boundonw}) is bounded as:
\[
\frac{1}{\lambda n} |{X}_j^\top (\vec{I}-\vec{X}_K(\vec{X}_K^\top \vec{X}_K)^{-1} \vec{X}_K^\top) \vec{u}|
\leq \frac{1}{\lambda n} \|X_j^\top (\vec{I}-\vec{X}_K(\vec{X}_K^\top \vec{X}_K)^{-1} \vec{X}_K^\top)\|_2  \cdot \|u\|_2 \leq \frac{\|X_j\|_2}{\lambda n} \cdot \|u\|_2.
\]
where we used the fact that $\vec{I}-\vec{X}_K(\vec{X}_K^\top \vec{X}_K)^{-1} \vec{X}_K^\top$ is an orthogonal projection matrix. By Assumption \ref{ass:id}, $\|X_j\|_2 \leq K\sqrt{n}$. It remains to bound $\|u\|_2$.

The first term is bounded by $1-\beta$ by Assumption \ref{ass:inc}. For the second term, note that $\|u\|=O(1)$ because they correspond to truncations according to sets that are close as we have already shown that $\|\check{\vec{\Omega}}-\vec{\Omega}^*\|\leq r$. Our desired bound follows by choice of $\lambda$.}
\end{proof}

%% file: future.tex

\section{Experimental Evaluation and Conclusion}

We studied the problem of parameter estimation for sparse Gaussian Graphical models and the problem of sparse linear regression, given samples that are subject to truncation. We provided sample efficient estimators for both aforementioned problems under suitable	 assumptions. 

We conducted a few experiments to empirically investigate the problem of inferring Gaussian graphical models. The algorithm we used was a projected stochastic gradient descent algorithm. In each iteration of this algorithm, the current estimates $\vec{v}$ and $\vec{\Theta}$ are updated by adding a subgradient of the graphical Lasso objective (\ref{eq:newfunction}), scaled by a regularization parameter (that is set in accordance with Lemma \ref{lem:mainbig}). The updated $\vec{\Theta}$ is projected so as to ensure that it is symmetric with minimum eigenvalue at least $10^{-5}$.
The distribution generating the original samples is a 10-dimensional Gaussian distribution with each co-ordinate truncated on a support  $\left(-2,2\right)$. The mean of the distribution $\vec{v}^*$ is set to be $\left(0,0,\cdots,0\right)$. Moreover, we set the precision matrix $\vec{\Theta}^*$ to be the identity matrix plus $0.2$'s entries in the the upper and lower diagonal, thus making the number of nonzero entries in the precision matrix to be $30$ (out of $100$).

\begin{figure}[t]
\centering
\begin{subfigure}{.5\textwidth}
  \centering
  \includegraphics[width=1.0\linewidth]{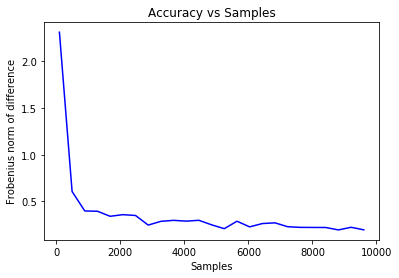}
  \caption{Error vs Number of Samples}
  \label{fig:samples}
\end{subfigure}%
\begin{subfigure}{.5\textwidth}
  \centering
  \includegraphics[width=1.0\linewidth]{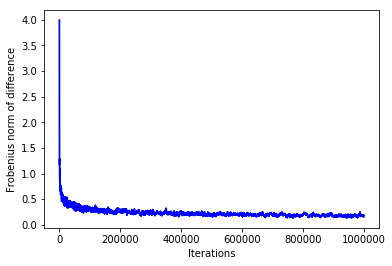}
  \caption{Error vs Number of Iterations of SGD}
  \label{fig:iterations}
\end{subfigure}
\end{figure}

The first experiment studies how the Frobenius norm error $\epsilon$ (between the true parameters and the estimates) varies with the number of samples  (fixing the number of iterations of SGD to be $10^6$).  The second experiment focuses on how the error varies with number of iterations of projected SGD for a fixed number of samples from the true distribution. 
In Figure \ref{fig:samples} we see that the number of samples scales, as expected, like $1/\epsilon^2$ w.r.t the error $\epsilon$. Figure \ref{fig:iterations} shows that projected SGD performs rather poorly computationally; this is also expected as the function we optimize $L_n$ is locally strong convex and the initialization is not necessarily close enough to the true parameters. 

We also performed a couple of experiments to understand how well we can recover the support of the model. One experiment studies how the sparsity of the estimate varies as a function of the number of iterations of the proposed algorithms for a fixed number of samples from the true distribution. Another focuses on understanding how the sparsity of the estimate varies with the number of initial samples (fixing the number of SGD iterations to $10^6$). To quantify the sparsity in our solution, we  “binarized" the estimated precision matrix by thresholding all elements
lying in the interval $[-0.1,0.1]$ to zero and the rest to one.

Now to compare the ``closeness" in sparsity between the two binary matrices (the original and the estimated), we evaluate element wise Hamming distance between them. This is defined as follows:
\begin{equation}
h(\vec{x},\vec{y})=\sum_{i=1}^{k}\mathbf{1}\{x_i \neq y_i\},
\end{equation}
where $\vec{x}$ and $\vec{y}$ are two $k$-dimensional binary vectors.

\begin{figure}[h]
	\centering
	\begin{subfigure}{.5\textwidth}
		\centering
		\includegraphics[width=1.0\linewidth]{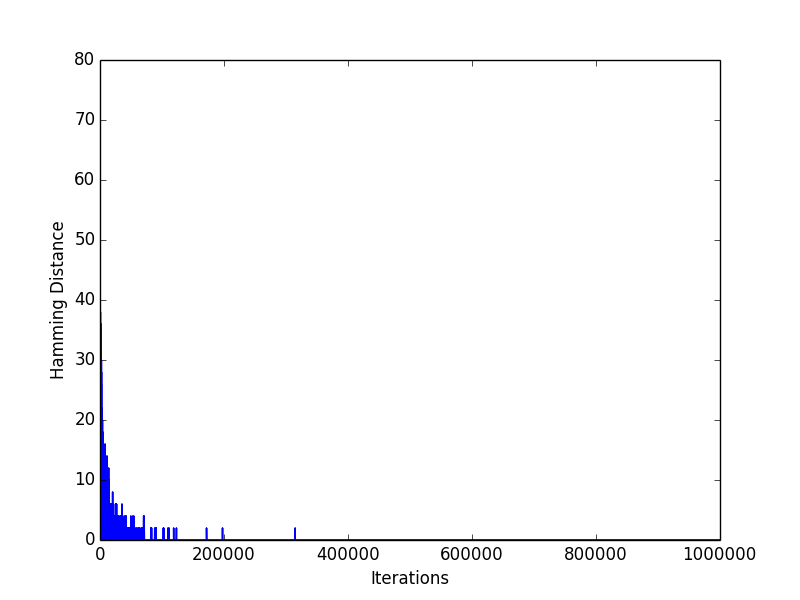}
		\caption{Hamming distance vs Iterations($n=50000$)}
		\label{fig:ham_iter50}
	\end{subfigure}%
	\begin{subfigure}{.5\textwidth}
		\centering
		\includegraphics[width=1.0\linewidth]{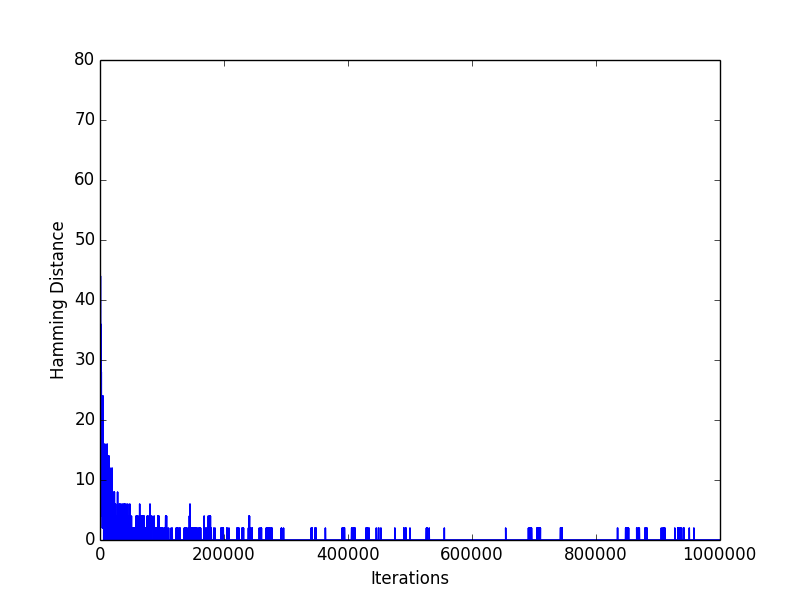}
		\caption{Hamming distance vs Iterations($n=5000$)}
		\label{fig:ham_iter5}
	\end{subfigure}
     \caption{Hamming distance vs Iterations in low and high sample regimes.}
     \label{fig:hamvsiter}
\end{figure}

\begin{figure}
	\centering
	\includegraphics[width=0.5\linewidth]{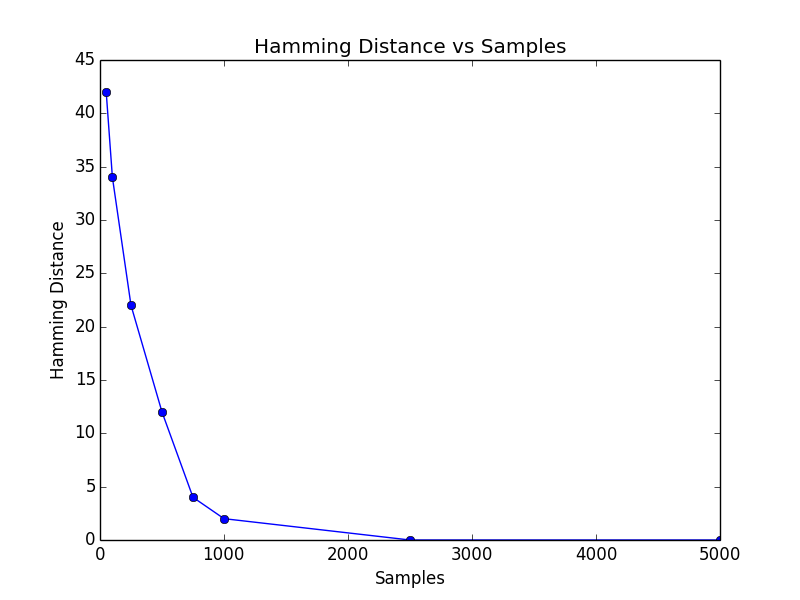}
	 \caption{Hamming distance vs Number of Samples when keeping the number of iterations to be $10^6$}
	 \label{fig:ham_samples}
\end{figure}

\paragraph{Evaluation.} The experiments suggests that our proposed algorithms recovers the true sparsity of the precision matrix. We observe from Figure \ref{fig:ham_iter50} that the sparsity of the estimate goes reaches the required number over $2*10^5$ number of iterations. Although the estimate is sparse, we observe from our previous experiments (Figure \ref{fig:iterations}), that the desired accuracy is achieved with ($\approx 10^6$) number of iterations.
In addition, our experiment for $n=5000$ (see Figure \ref{fig:ham_iter5}) suggests that even if we obtain a sparse estimate with less than $50000$ initial samples, it takes a larger number of iterations to converge. This is evidenced in Figure \ref{fig:ham_iter5}, where, the graph is noisy before ($\approx 10^6$) iterations and convergence happens beyond it.

In our second experiment we try to understand how the hamming distance varies with the initial number of samples. From Figure \ref{fig:ham_samples}, we observe that $n=2500$ are sufficient to achieve the required sparsity with $10^6$ iterations. 

Thus, together with the experiments on the convergence with respect to the Frobenius norm of the precision matrix, we can see that our proposed algorithm is able to recover the sparse precision matrix with high accuracy.

One future direction is the computational efficiency, i.e., to come up with polynomial time algorithms that compute estimators for the abovementioned problems with same sample complexity guarantees as in our claims. 
Another is to provide finite sample bounds for recovering the support of a Gaussian graphical model. The assumptions we made for support recovery of sparse linear models in Theorem \ref{thm:main2} are too strong for us to invoke the ``neighborhood selection approach'' of \cite{Mein06}.

%% file: ell2.tex

\section{Bounds on $\ell_2$-error for truncated linear regression}\label{sec:ell2}

In this section, we give bounds on the $\ell_2$-error for the truncated sparse linear regression problem. The proofs are straightforward given the techniques we have used elsewhere in this paper.

We use the same notation as Section \ref{sec:linreg}, with:
\begin{equation}\label{eqn:minhat}
\hat{\vec{\Omega}} \in \arg \min_{\vec{\Omega} \in \R^d} \{l_n(\vec{\Omega})+ \lambda \|\vec{\Omega}\|_1\}
\end{equation}
where $l_n$ is the negative log-likelihood function:
\[
l_n(\vec{\Omega}) = \frac1n \sum_{i=1}^n \left(\frac{1}{2} {y^{(i)}}^2 - y^{(i)} \vec{\Omega}^\top \vec{x}^{(i)}  + \log \int \exp\left(-\frac{1}{2} z^2 +z \vec{\Omega}^\top \vec{x}^{(i)}\right) S(z) dz\right),
\]
which is convex everywhere. Recall that each $y^{(i)}$ is distributed as $\mathcal{N}(\vec{x}^{(i)}\vec{\Omega}^*, 1, S)$; we denote $K$ as the support of $\vec{\Omega}^*$ and let $k = |K|$. The matrix $\vec{X}$ is defined to have $n$ rows $\vec{x}^{(1)}, \dots, \vec{x}^{(n)}$. 

For the theorem in this section, we need the survival probability assumption (Assumption \ref{ass:sur}) and the normalization assumption (Assumption \ref{ass:norm}), but we replace the minimum eigenvalue and mutual incoherence assumptions by the following.

\begin{assumption}[Restricted Eigenvalue]\label{ass:reseig}
There exists $\beta > 0$ such that the matrix $\vec{X}$ satisfies:
\[\frac{1}{n} \|\vec{X} \vec{\Delta}\|_2^2 \geq \beta \|\vec{\Delta}\|_2^2
\quad\quad \text{for all } \vec{\Delta} \text{ such that } \|\vec{\Delta}_{\overline{K}}\|_1 \leq 3 \|\vec{\Delta}_K\|_1.
\] 
\end{assumption}
The restricted eigenvalue assumption is a very common one in the study of Lasso-type algorithms, and it is known that 
many families of random design matrices satisfy it (see Chapter 7 of \cite{Wbook}). We can now state the main result of this section:

\begin{theorem}
Suppose that Assumptions \ref{ass:sur}, \ref{ass:norm} and \ref{ass:reseig} hold. Then, for any $\eps > 0$, if $\lambda = O\left(\frac{\beta \alpha^4 \eps}{Ck}\right)$ and $n = \tilde{O}\left(\frac{C^4k^2 \log d}{\beta^2 \alpha^8\eps^2}\right)$, then $\|\hat{\vec{\Omega}} - \vec{\Omega}^*\|_2 \leq \eps$. 
\end{theorem}
\begin{proof}
We begin with a lemma showing ``restricted strong convexity'' of $l_n$ in a neighborhood of $\vec{\Omega}^*$. 
\begin{lemma}\label{lem:locresconv}
There exists $\kappa \geq \Omega(\alpha^4\beta)$ for which the following holds. For any $\vec{\Omega}$ such that $\vec{\Delta} = \vec{\Omega}-\vec{\Omega}^*$ satisfies: (i) $\|\vec{\Delta} \|_2 \leq \frac{1}{C\sqrt{k}}$ and (ii) $\|\vec{\Delta}_{\overline{K}}\|_1 \leq 3 \|\vec{\Delta}_K\|_1$, it holds that:
 \[\vec{\Delta}^\top \left(\nabla^2 l_n(\vec{\Omega})\right)\vec{\Delta} \geq \kappa \|\vec{\Delta}\|_2^2.\]
\end{lemma}
\begin{proof}
Using the same steps as in the proof of Lemma \ref{lem:locstrong}, we get:
	\begin{align*}
\vec{\Delta}^\top \left(\nabla^2 l_n(\vec{\Omega})\right)\vec{\Delta} 
		\geq \frac{1}{12e^2n} \sum_{i=1}^n ({\vec{x}^{(i)}}\vec{\Delta})^2 (\alpha(\vec{\Omega}^*, \vec{x}^{(i)}))^4 \cdot \exp\left(-2|\vec{x}^{(i)}(\vec{\Omega}-{\vec{\Omega}^*})|^2-4\right)
	\end{align*}
	Now, $|\vec{x}^{(i)}\vec{\Delta}|\leq \|\vec{x}^{(i)}\|_\infty \|\vec{\Delta}\|_1$. From Assumption \ref{ass:norm}, $\|\vec{x}^{(i)}\|_\infty \leq C$, while from the lemma's conditions, $\|\vec{\Delta}\|_1 = \|\vec{\Delta}_K\|_1 + \|\vec{\Delta}_{\overline{K}}\|_1 \leq 4 \|\vec{\Delta}_K\|_1 \leq 4 \sqrt{k}\|\vec{\Delta}\|_2 \leq 4/C$. Therefore:
	\begin{align*}
	\vec{\Delta}^\top \left(\nabla^2 l_n(\vec{\Omega})\right)\vec{\Delta} 
		\geq \frac{\alpha^4}{12e^{38}n} \|\vec{X}\vec{\Delta}\|_2^2 \geq \frac{\alpha^4\beta}{12e^{38}}\|\vec{\Delta}\|_2^2,
		\end{align*}
		where the last inequality uses Assumption \ref{ass:reseig}.
\end{proof}

In the following, let $\vec{\Delta} = \hat{\vec{\Omega}} - \vec{\Omega}^*$ where $\hat{\vec{\Omega}}$ is as defined in (\ref{eqn:minhat}). The next lemma is well-known and uses only the convexity of $l_n$.
\begin{lemma}[e.g., Proposition 9.13 of \cite{Wbook}]\label{lem:wainwright913}
If $\lambda \geq 2 \|\nabla l_n(\vec{\Omega}^*)\|_\infty$, then $\|\vec{\Delta}_{\overline{K}}\|_1 \leq 3 \|\vec{\Delta}_K\|_1$.
\end{lemma}
Combining the above, we can now follow the proof of Lemma \ref{lem:mainbig} to get a similar statement.
\begin{lemma}
Suppose $\lambda \geq 2 \|\nabla l_n(\vec{\Omega}^*\|_\infty$. For any $\eps > 0$, if $\lambda = O\left(\frac{\beta \alpha^4 \eps}{Ck}\right)$, then $\|\Delta\|_2 \leq \eps$.
\end{lemma}
\begin{proof}
We first argue that if $\|\vec{\Delta}\|_2 > \frac{1}{C\sqrt{k}}$:
\[
|\vec{\Delta}^\top (\nabla l_n(\hat{\vec{\Omega}}) - \nabla l_n(\vec{\Omega}^*))| \geq \Omega\left(\frac{\beta\alpha^4}{C\sqrt{k}}\right) \|\vec{\Delta}\|_2.
\]
This follows by combining Lemma\footnote{We actually need an extension of Lemma \ref{lem:wainwright910} where $\vec{\Delta}$ is restricted to the cone $\mathbb{C} = \{\vec{z}: \|\vec{z}_{\overline{K}}\|_1 \leq 3 \|\vec{z}_K\|_1$\}. The proof goes through unchanged.} \ref{lem:wainwright910}, Lemma \ref{lem:locresconv} and Lemma \ref{lem:wainwright913}. We also have the upper bound as in the proof of Lemma \ref{lem:mainbig}:
\[
|\vec{\Delta}^\top (\nabla l_n(\hat{\vec{\Omega}}) - \nabla l_n(\vec{\Omega}^*))| \leq 
\|\vec{\Delta}\|_1 \left(\lambda + \|\nabla l_n(\vec{\Omega}^*)\|_\infty\right) \leq \frac{3 \lambda}{2} \|\vec{\Delta}\|_1 \leq 6\sqrt{k}\lambda \cdot \|\vec{\Delta}\|_2
\]
where we used the lower bound on $\lambda$ and the fact $\|\vec{\Delta}\|_1 \leq 4\sqrt{k}\|\vec{\Delta}\|_2$ due to Lemma \ref{lem:wainwright913}. The above two inequalities contradict if $\lambda =O\left(\frac{\beta \alpha^4}{Ck}\right)$ and hence, $\|\vec{\Delta}\|_2 \leq \frac{1}{C\sqrt{k}}$. 

Continuing along the lines of Lemma \ref{lem:mainbig}, we see that:
\[
|\vec{\Delta}^\top (\nabla l_n(\hat{\vec{\Omega}}) - \nabla l_n(\vec{\Omega}^*))| \geq  \Omega(\beta \alpha^4) \|\vec{\Delta}\|_2^2
\]
Hence, we again get a contradiction to $\|\vec{\Delta}\|_2 > \eps$ if $\lambda = O\left(\frac{\beta \alpha^4 \eps}{Ck}\right)$.
\end{proof}
Finally, we need an upper bound to $\|\nabla l_n(\vec{\Omega}^*\|_\infty$. We already carried out essentially this calculation in the proof of Lemma \ref{lem:lambda}. It follows from there that $\|\nabla l_n(\vec{\Omega}^*\|_\infty < t$ with high probability if $n \geq \Omega(C^2 \log d \log(n/\alpha)/t^2)$. Setting $t = \lambda/3 = O\left(\frac{\beta \alpha^4\eps}{Ck}\right)$, we get that it's sufficient to set $n \geq \tilde{\Omega}\left(\frac{C^4k^2 \log d}{\beta^2 \alpha^8\eps^2}\right)$.

\end{proof}